\RequirePackage{tikz}

\documentclass[sn-mathphys]{sn-jnl}

\usepackage{algorithm}
\usepackage{verbatim}
\usepackage{multirow}
\usepackage{subfig}
\usepackage{graphicx}
\usepackage{rotating}
\usepackage{csquotes}
\usepackage{physics}
\usepackage[title]{appendix}
\usepackage{program}
\usepackage{float}
\usepackage{tikz}
\usetikzlibrary{shapes, positioning}
\usepackage{pgfplots}
\usepgfplotslibrary{statistics}
\pgfplotsset{compat=1.16}

\jyear{2022}%

\theoremstyle{thmstyleone}%
\theoremstyle{thmstyletwo}%

\theoremstyle{thmstylethree}%

\newcommand{\mat}[1]{{\ensuremath{{\mathbf{#1}}}}}
\def\A{\mat A}
\def\E{\mat E}
\def\Q{\mat Q}
\def\T{^\mathrm{T}} 
\def\DEF{\triangleq} 

\begin{document}

\title[Explaining Classification Data Sets with Decision Rules]{A Nested Genetic Algorithm for Explaining Classification Data Sets with Decision Rules}


\author{Paul-Amaury Matt$^1$}\email{paul-amaury.matt@ipa.fraunhofer.de}

\author{Rosina Ziegler$^1$}\email{rosina.ziegler@ipa.fraunhofer.de}

\author{Danilo Brajovic$^1$}\email{danilo.brajovic@ipa.fraunhofer.de}

\author{Marco Roth$^1$}\email{marco.roth@ipa.fraunhofer.de}

\author{Marco F. Huber$^{1,2}$}\email{marco.huber@ieee.org}

\affil{$^1$Department of Cyber Cognitive Intelligence (CCI), Fraunhofer Institute for Manufacturing Engineering and Automation IPA, Nobelstrasse 12, 70569 Stuttgart, Germany\\
\\
$^2$Institute of Industrial Manufacturing and Management IFF, University of Stuttgart, Allmandring 35, 70560 Stuttgart, Germany}


\abstract{Our goal in this paper is to automatically extract a set of decision rules (rule set) that best explains a classification data set. First, a large set of decision rules is extracted from a set of decision trees trained on the data set. The rule set should be concise, accurate, have a maximum coverage and minimum number of inconsistencies. This problem can be formalized as a modified version of the weighted budgeted maximum coverage problem, known to be NP-hard. To solve the combinatorial optimization problem efficiently, we introduce a nested genetic algorithm which we then use to derive explanations for ten public data sets.}




\maketitle

\section{Introduction}
\label{section-introduction}

In 1959, Arthur Samuel, a pioneer in the field of Artificial Intelligence defined  the term \emph{Machine Learning} \cite{samuel1959some} as the \enquote{field of study that gives computers the ability to learn without being explicitly programmed}. In the field of Machine Learning, an important technique called \emph{Deep Learning} allows \enquote{computational models that are composed of multiple processing layers to learn representations of data with multiple levels of abstraction} \cite{lecun2015deep}. In recent years, many accurate decision support systems based on Deep Learning have been constructed as \emph{black boxes} \cite{guidotti2018survey}, that is as systems that hide their internal logic to the user.

Thus, the purpose of an \emph{Explainable Artificial Intelligence} \cite{molnar2020interpretable, samek2019explainable, xu2019explainable, burkart2021survey} system is to make its behavior more intelligible to humans by providing explanations \cite{gunning2019xai}. A popular approach to addressing the problem of opacity of black-box machine learning models is the use of \emph{post-hoc} explainability methods: these methods approximate the logic of underlying machine learning models with the aim of explaining their internal workings, so that the user can understand them \cite{vale2022explainable}. 

Unfortunately, these methods provide explanations that are not faithful to what the black-box model computes and can be misleading \cite{rudin2019stop}. A recent and highly cited perspective \cite{rudin2019stop} highlighted the need for \emph{white box} models (i.e. models explainable by nature) for high stakes decisions in Healthcare and Criminal Justice. Other application domains of Artificial Intelligence where white box models are deemed to be important are Banking, Bioinformatics, Automotive, Marketing, Politics, Agriculture, Defense and Recommender Systems \cite{burkart2021survey}. Examples of white box models are linear regression, decision trees, mixtures of decision trees \cite{vasic2019moet} and rule sets \cite{wang2017bayesian}.

\begin{figure}[h!]
\begin{center}
    \includegraphics[width=11.5cm]{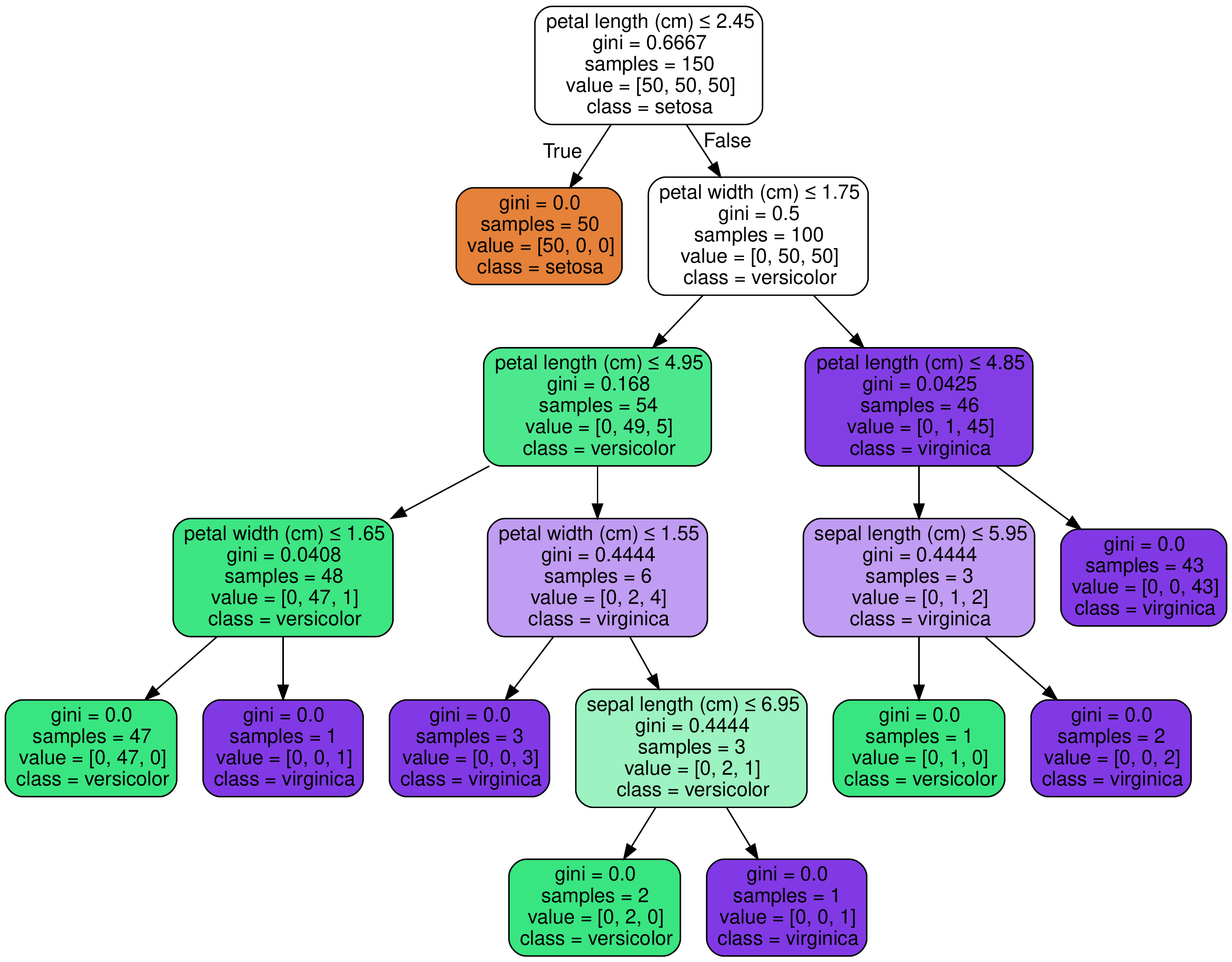}
\end{center}
\caption{Decision tree trained on the Iris data set.}
\label{figure:iris-decision-tree}
\end{figure}

In this article, we focus on an important Machine Learning task, namely \emph{classification}. Formally, classification refers to a predictive modeling problem where a class label is predicted for a given example of input data. For instance, given the length and width in centimeters of the petals and sepals of an iris flower, we want to determine if the flower is from the setosa, versicolour or virginica species. The Iris data set\footnote{\url{https://archive.ics.uci.edu/ml/datasets/iris}} contains 150 instances of flowers, with 50 instances for each of these 3 species. Using this data set, we could build classifiers using classification models such as decision trees, random forests, support vector machines and neural networks. But amongst these, only a sufficiently small decision tree would yield an \enquote{explainable} solution. Let us for instance consider the decision tree shown in Figure \ref{figure:iris-decision-tree} that has been obtained with Python's scikit-learn\footnote{\url{https://scikit-learn.org/stable/modules/tree.html}}. 

The decision tree can be transformed into a set of decision rules or so-called \emph{rule set} in the following way. Every path starting at the root node and ending at one of the leaf nodes is turned into a decision rule. A \emph{decision rule} is an IF-THEN statement consisting of a conjunction of conditions followed by a conclusion in the form of a class prediction: \\[0.2cm]
\emph{IF} condition-1 \emph{AND} ... \emph{AND} condition-n \emph{THEN} CLASS=y \\[0.2cm]
The internal nodes and branches followed along the path make up the conditions and the leaf node's majority class makes up the predicted class. Decision rules follow a general structure: IF all the conditions are met THEN make a certain prediction. Decision rules are probably the most interpretable prediction models. Their IF-THEN structure semantically resembles natural language and the way we think, provided that the conditions are built from intelligible features of the input data and that the explanation is sufficiently concise. In the example of Figure \ref{figure:iris-decision-tree}, there are 9 leaf nodes and thus also 9 decision rules that can be extracted from the decision tree. See Figure \ref{figure:iris-decision-tree-rules}.

\begin{figure}[h!]
\begin{tiny}
\begin{verbatim}
IF petal length<=2.45 THEN CLASS=setosa
IF petal length>2.45 AND petal width<=1.75 AND petal length<=4.95 AND petal width<=1.65 THEN CLASS=versicolor
IF petal length>2.45 AND petal width<=1.75 AND petal length<=4.95 AND petal width>1.65 THEN CLASS=virginica
IF petal length>2.45 AND petal width<=1.75 AND petal length>4.95 AND petal width<=1.55 THEN CLASS=virginica
IF petal length>2.45 AND petal width<=1.75 AND petal length>4.95 AND petal width>1.55 AND sepal length<=6.95 
THEN CLASS=versicolor
IF petal length>2.45 AND petal width<=1.75 AND petal length>4.95 AND petal width>1.55 AND sepal length>6.95 
THEN CLASS=virginica
IF petal length>2.45 AND petal width>1.75 AND petal length<=4.85 AND sepal length<=5.95 THEN CLASS=versicolor
IF petal length>2.45 AND petal width>1.75 AND petal length<=4.85 AND sepal length>5.95 THEN CLASS=versicolor
IF petal length>2.45 AND petal width>1.75 AND petal length>4.85 THEN CLASS=virginica

\end{verbatim}
\end{tiny}
\caption{Decision rules derived from the decision tree in Figure \ref{figure:iris-decision-tree}.}
\label{figure:iris-decision-tree-rules}
\end{figure}

In the example, the decision rules give a prediction for all instances of the data set and all predictions made are correct. In general, the decision rules obtained from a decision tree never conflict, because for any data input there can by only one decision rule that applies. However, the decision rules obtained from a decision tree may lead to prediction errors.

The complexity or \emph{length} of a decision rule can be simply defined by its number of conditions. Here, the simplest rule is the first one and has a length of 1. The most complex rules (fifth and sixth rules) have a length equal to the depth of the decision tree, which is 5. The sum of the lengths of all decision rules can be used as a measure of the complexity of the explanation for the data set. In our example, the explanation has a complexity of 34. 

Is there a simpler explanation of the Iris data set? To check if that is the case, we could start by training several decision trees on the data set, for instance by training a random forest or by using a decision tree algorithm with different hyper-parameter combinations, varying for example the maximum tree depth or information criterion. Each tree would yield its own set of decision rules and we would obtain explanations that differ in terms of complexity. But the simpler the explanation, the larger the number of prediction errors would potentially be. So we would have to set a maximum number of prediction errors as constraint to answer the question in a meaningful way.

An even more general idea would be to gather all the decision rules from all the trees generated and to select rules from that large set. A set of decision rules selected in this way may not necessarily predict all instances, unlike those obtained from decision trees. One should also expect to encounter conflicts between the rules, but this more general method would allow us to find simpler explanations and reduce the number of classification errors.

Several areas of Artificial Intelligence may benefit from an automated procedure for explaining a data set with decision rules, such as 1) Data Science, where one essentially seeks to derive insights from data, 2) Knowledge Representation, where one may be interested in automatically extracting from data some knowledge expressed in a logical form, 3) Formal Specification, as one may replace the classes of the data set by the predicted classes of a black-box model and by extracting rules get a specification of the behaviour of the model, and finally 4) Explainable Artificial Intelligence, where one seeks to build models that are \enquote{transparent}, i.e. that can be explained.

The remainder of the article is organised as follows. In Section \ref{section-related-work}, we discuss existing approaches for rule extraction and other related works. In Section \ref{section-generation}, we examine how a set of candidate decision rules may be generated. Section \ref{section-formulation} formalizes the problem of finding an optimal explanation as a budgeted maximum coverage problem \cite{khuller1999budgeted}, known to be NP-hard. A \emph{genetic algorithm} (GA) is an iterative search, optimization and adaptive machine learning technique premised on the principles of natural selection. Genetic algorithms are capable of finding solutions to NP-hard problems \cite{panchal2015solving}. In Section \ref{section-nestedGA}, we introduce a nested genetic algorithm for solving the rule selection problem, whereby the inner genetic algorithms are necessary for facilitating the search in the constrained search space and enable the evolutionary process. In Section \ref{section-experiments}, we then test the performance of the algorithm on several simple classification data sets from the UCI Machine Learning Repository and compare its performance with adaptations of the methods exposed in Section \ref{section-related-work} and conclude in Section \ref{section-conclusion}.

In the remainder of the paper, we denote $(X,Y)$ the classification data set, where $X=x_1,\dotsc,x_N$ are $N$ vectors of dimension $d$ and $Y=y_1, \dotsc, y_N$ are labels that belong to a finite set of classes $C$. We denote $R_1, \dotsc, R_M$ the decision rules generated and $M$ their number.

\section{Related Work}
\label{section-related-work}

This section reviews some of the main approaches existing for rule extraction from an ensemble of decision trees. The approaches exposed in this section aim at increasing the explainability of the ensemble and its predictive accuracy. Recall that strictly speaking, a rule set does not constitute an ensemble per se. But, a rule set allows to build a classifier once we specify a voting mechanism to elect a winning class. Most often, the output class of the ensemble is determined by weighted majority voting and the accuracy of the ensemble classifier can then be defined as the fraction of observations correctly classified. Our approach differs from the existing works discussed here-after, since a) we do not explicitly seek to maximize \emph{accuracy} (fraction of correctly classified data) but rather the \emph{coverage} (fraction of data classified unambiguously by the rules) of the solution, and b) these approaches do not impose any hard constraints on the level of complexity of the explanation nor on the number of classification errors. 

\subsection{ForEx++}

The ForEx++ algorithm \cite{AdnanIslam2017forex} starts with the set of rules obtained from a decision forest and removes identical rules. For each class, it computes three sets of rules: 1) the set of rules with accuracy above average, 2) the set of rules with coverage above average and 3) the set of rules with length below average. The intersection of these three sets is then computed for each class. Finally, the algorithm returns the union over all classes of these intersections. The rules extracted by this method are comparatively more accurate, general and concise than others. Some drawbacks of the method are that the rules may not be very diverse (similar rules may be selected for each class) and they may not cover the entire data set.

\subsection{RF+HC Method}
    
The RF+HC algorithm \cite{gras2015rfhc} is a rule extraction method to extract comprehensible rules from a random forest. The algorithm consists of four parts. First, the rules are generated. Then, a score is computed for each rule according to
$$ \mathrm{ruleScore} = \frac{cc - ic}{cc + ic} + \frac{cc}{ic+k} + \frac{cc}{rl}$$
where $cc$ is the number of observations that are covered and correctly classified by the rule, $ic$ is the number of incorrectly classified observations that are covered by the rule, $rl$ is the rule length and $k$ is a positive constant (set to $k=4$ by the authors). This scoring function ensures the retention of rules with the highest levels of classification accuracy and the highest levels of coverage, along with prioritizing concise rules. In this \enquote{hill climbing} approach (HC), the rule set contains initially all rules. Rules are then iteratively removed from the set. In each iteration, a rule is probabilistically selected on the basis of its score (meaning that a rule with low score has greater chance to be selected than a rule with high score). If the removal of the selected rule increases the overall accuracy, then the change is retained. The removal process terminates when all rules have been considered. Since the outcome is random, the process is repeated several times and the rule set that achieves the highest accuracy is returned by the algorithm.

\subsection{IRFRE}

The Improved Random Forest based Rule Extraction (IRFRE) method \cite{wang2020irfre} derives accurate and interpretable classification rules from a decision tree ensemble. Decision rules are extracted from a Random Forest and then selected using a multi-objective genetic algorithm \cite{fonseca1993multiobjective}. An initial population is randomly generated, where each individual's chromosome is coded as a binary string $r_1 \dotsc r_M$ of length $M$ (the number of rules generated) and $r_i=1$ if rule $R_i$ is selected and $r_i=0$ otherwise. For every generation, the accuracy and interpretability of each individual (rule set) are calculated. The interpretability of the classifier is defined as a linear combination of the average length of the rules, the number of rules and their coverage. The rank of each individual is determined and a randomized mating process takes place, in which individuals with lowest rank are selected with highest probability. Each pair of individuals gives birth to two children after uniform crossover and bitflip mutation. The best individuals amongst parents and children of the current generation are selected for the next generation. When the maximum number of generation is reached, the algorithm stops. This yields a final population and its Pareto optimal solutions are returned.

\subsection{Other Related Works}

Heuristic methods are traditionally used to quickly produce decision trees with reasonably high accuracy. A commonly criticised point, however, is that the resulting trees may not be the best representation of the data in terms of accuracy and size. \cite{izza2020explaining, izza2022tackling} show that in some settings decision trees can hardly be deemed interpretable, with paths being arbitrarily larger than a minimal set of features that entails the prediction (i.e., a so-called \emph{PI-explanation}).

This motivated the development of optimal classification tree algorithms that globally optimise the decision tree in contrast to heuristic methods that perform a sequence of locally optimal decisions. \cite{verhaeghe2020learning} introduced an approach to learn decision trees using constraint programming that allows to limit the size of the decision trees while maintaining a good classification accuracy. \cite{alos2021learning} offers a combinatorial optimization approach based on Maximum Satisfiability technology to compute Minimum Pure Decision Trees for the sake of interpretability. These decision trees can outperform on average the decision tree classifiers generated with Python's scikit-learn in terms of accuracy. \cite{demirovic2022murtree} provides a novel algorithm for learning optimal classification trees based on dynamic programming and search that supports constraints on the depth of the tree and number of nodes.

\cite{lakkaraju2016interpretable} formalizes decision set learning through an objective function that simultaneously optimizes for accuracy, conciseness, and meaningfulness of the rules. Their approach learns short, accurate and non-overlapping
rules that cover the whole feature space and pay attention to small
but important classes. Their experiments show that interpretable decision sets are as accurate at classification as state-of-the-art machine learning techniques. Moreover, interpretable decision sets are three times smaller than models learned by competing approaches. \cite{wang2017bayesian} present a machine learing algorithm for building probabilistic models for classification that are comprised of a small number of short rules in disjunctive normal form, which have the advantage of being interpretable to human experts.

\section{Generation of Decision Rules}
\label{section-generation}

In this section, we briefly review how decision rules can be generated and specify how we do this in our implementation. In general, we may distinguish six steps for generating a set of decision rules from decision trees: 
\begin{enumerate}
    \item \textbf{Choice of a decision tree algorithm}. We may choose amongst several decision tree algorithms such as ID3, C4.5, C5.0 or CART. We use the Python package scikit-learn which implements an optimised version of the CART algorithm. Note however that this algorithm does not support categorical variables, so the categorical variables of the data set need to be one-hot encoded before using the algorithm.
    \item \textbf{Data preparation}. Prepare the data so that the chosen training algorithm can run without error. For instance, we may have to remove entries with missing values or fill the missing values by applying an imputation technique. Transform each categorical variable into several binary variables by one-hot encoding, if the chosen algorithm cannot handle categorical variables. Optionally, split randomly the available data into a training and a testing set if the user's ultimate goal is to build a classifier from the rule set.
    \item \textbf{Choice of hyper-parameter values}. Choose one or several combinations of values for the hyper-parameters of the algorithm. In our implementation, the hyper-parameter combinations are shown in Table \ref{tab:hyper-parameters-table}.
    \item \textbf{Train trees or random forests}. For each hyper-parameter combination, train a tree on the whole data set / training set, or train a random forest, whereby each tree of the forest is trained on a random sub-sample of the data. We train decision trees rather than random forests to minimize the number of rules generated.
    \item \textbf{Transformation into rules}. Each path from the root node to a leaf node of a decision tree is transformed into a decision rule. The internal nodes and branches followed along the path path yield the conditions of the decision rule and the majority class of the leaf node yields the predicted class.
    \item \textbf{Simplification}. We then simplify redundant conditions within a decision rule. For instance, two conditions on the same feature $k$ such as $x^k \geq \alpha$ and $x^k \geq \beta$ simplify as one condition $x^k \geq \max(\alpha, \beta)$. Remove rules that are redundant (identical) or empirically equivalent, i.e. that yield the same predictions on the data set. The reader interested in a more advanced method for rule simplification may refer to \cite{izza2020explaining}.
\end{enumerate}

\begin{table}
\centering
\caption{Hyper-parameters used for the CART algorithm.}
\label{tab:hyper-parameters-table}
\begin{tabular}{ |p{2cm}|p{5cm}|p{2cm}|  }
 \hline
 \multicolumn{3}{|c|}{Hyper-parameter} \\
 \hline
 Name & Definition & Range of values \\
 \hline
 \hline
 criterion & The function to measure the quality of a split. Supported criteria are “gini” for the Gini impurity and “entropy” for the information gain. & \{"gini", "entropy"\} \\
 \hline
 splitter & The strategy used to choose the split at each node. Supported strategies are “best” to choose the best split and “random” to choose the best random split. & \{"best", "random"\} \\
 \hline
 max\_depth & The maximum depth of the tree. & integer \\
 \hline
 max\_features & The number of features to consider when looking for the best split. If “sqrt”, then max\_features=$\sqrt{d}$. If “log2”, then max\_features=$\log_2(d)$. & \{"sqrt", "log2"\}\\
 \hline
\end{tabular}
\end{table}

It is also possible to generate decision rules by \emph{rule mining} \cite{wang2017bayesian}. For categorical attributes, consider both positive associations (e.g. $x_j=\mathrm{blue}$) and negative associations ($x_j \neq \mathrm{green}$). For each numerical attribute, create a set of binary variables by comparing it with a set of thresholds (usually quantiles) and add their negation as separate attributes. Then mine for frequent rules within the sets of observations pertaining to the different classes, using some frequent rule-mining method such as the FP-growth algorithm \cite{borgelt2005implementation}. The number of rules may be reduced by imposing a maximum length of rules, a minimum support or a minimum information gain.

\section{Formulation of the Optimisation Problem}
\label{section-formulation}

Let us now formalize the problem of selecting an optimal rule set for explaining a classification data set. Intuitively, the selected rules should achieve several objectives. Firstly, we would like the rule set to predict a class for a maximum number of instances in the data set. Secondly, the number of conflicts between the rules should be minimized. Thirdly, the total complexity of the rules shall not exceed a given complexity fixed by the user. Finally, the total number of classification errors should not exceed a given number of classification errors. We may consequently formulate the optimal rule selection problem as a modified budgeted maximum coverage problem \cite{khuller1999budgeted}:
\begin{align*}
\texttt{maximise } \sum_{j=1}^N e_j - \frac{\epsilon}{2} \sum_{j=1}^N \sum_{i=1}^M \sum_{i'=1}^M \alpha_{i,i',j} \cdot r_i \cdot r_{i'} \\
\texttt{subject to } L(r) = \sum_{i=1}^M L(R_i) \cdot r_i \leq B_c \\
E(r) = \sum_{i=1}^M E(R_i) \cdot r_i \leq B_e \\
\forall j \in \{ 1, \dotsc, N\}: \quad e_j \leq \sum_{i \in S_j } r_i \\
\forall i \in \{ 1, \dotsc, M\}: \quad r_i \in \{0,1\}\\
\forall j \in \{ 1, \dotsc, N\}: \quad e_j \in \{0,1\}\\
\end{align*}
where the decision variables and parameters are defined as follows:
\begin{itemize}
    \item $e_j$ is a binary variable equal to $1$ if and only if data input $x_j$ is covered by at least one of the selected rules
    \item $r_i$ is a binary variable equal to $1$ if and only if the decision rule $R_i$ is selected
    \item the binary coefficient $\alpha_{i,i',j}$ is equal to $1$ if and only if $R_i$ and $R_{i'}$ cover $x_j$ but lead to different predictions (conflict)
    \item $\epsilon \geq 0$ is the penalty for every conflict between a pair of selected rules
    \item $S_j$ is the set of indices of rules that cover $x_j$
    \item $L(R_i)$ is the length of $R_i$ 
    \item $E(R_i)$ is the number of observations incorrectly classified by $R_i$
    \item $B_c \geq 0$ is the budget for the total complexity of the rule set
    \item $B_e \geq 0$ is the budget for the total number of classification errors \footnote{In  practice, the user may determine the error budget by setting a fraction of error, multiplying the fraction by $N$ and rounding the result to the closest integer.}
\end{itemize}

The problem's objective function is quadratic and the inequality constraints are linear with respect to the decision variables $r_i$ and $e_j$. This combinatorial optimization problem involves $M+N$ binary variables,  where $M$ is the number of generated rules and $N$ is the number of data instances. It is fortunately possible to reduce the problem to $M$ variables. Indeed, every variable $e_j$ should be maximized, whilst satisfying $e_j \leq 1$ and $e_j \leq \sum_{i \in S_j} r_i$. So, for an optimal solution $r$, this implies $e_j = 1$ if $\sum_{i \in S_j} r_i \geq 1$ and $e_j=0$ otherwise. We thus introduce the functions $e_j^*: \{ 0,1\}^M \rightarrow \{ 0,1 \}$ for all $j \in \{ 1, \dotsc, N\}$:
$$ e_j^*(r) = \begin{cases}
    1 & \text{if } \sum_{i \in S_j} r_i \geq 1 \\
    0              & \text{otherwise}
\end{cases} $$
We may now get rid the $N$ variables $e_j$, thus ending up with an equivalent problem that requires only $M$ variables:
\begin{align*}
\texttt{maximise } f^*(r) = \sum_{j=1}^N e_j^*(r) - \frac{\epsilon}{2} \sum_{j=1}^N \sum_{i=1}^M \sum_{i'=1}^M \alpha_{i,i',j} \cdot r_i \cdot r_{i'} \\
\texttt{subject to } L(r) = \sum_{i=1}^M L(R_i) \cdot r_i \leq B_c \\
E(r) = \sum_{i=1}^M E(R_i) \cdot r_i \leq B_e \\
\forall i \in \{ 1, \dotsc, M\}: \quad r_i \in \{0,1\}\\
\end{align*}
Note that the objective function is not quadratic anymore. In the remainder of the article, we refer to the value of objective function $f^*(r)$ simply as coverage, the fraction $f^*(r)/N$ as coverage score, $L(r)$ is refered to as complexity of the rule set and $E(r)$ as the number of classification errors.

\section{Nested Genetic Algorithm}
\label{section-nestedGA}

To solve the problem above, we develop a nested genetic algorithm. Our algorithm uses ideas stemming from some of the classical approaches for rule extraction presented in Section \ref{section-related-work}, viz. the RF+HC and IRFRE methods. Since these two methods were created in a different context, we start this section by adapting them to fit our problem. We then introduce a genetic algorithm to solve the budgeted maximum coverage problem. The adapted RF+HC method is used to compute part of the initial population. We then modify the classical genetic algorithm and enhance it using two \enquote{inner} genetic algorithms nested in the first or \enquote{outer} algorithm. The modified IRFRE method will mainly be used to demonstrate experimentally that the inner genetic algorithms are necessary to enable the evolutionary process (crossover and genetic mutations) when searching for solutions in the constrained space.

\subsection{Adaptations of Existing Rule Extraction Methods}

In this subsection, we adapt three classical rule extraction methods to the budgeted maximum coverage problem.

\subsubsection{ForEx++}

The original ForEx++ Algorithm \ref{alg:forex} can be used without any modification for rule selection in our setting.

\begin{algorithm}[h]
\caption{ForEx++}
\begin{algorithmic}[1]
\For{each class $c$ in $C$}
\State let $\mathcal{I}_c$ be the intersection of the sets of the rules with 1) accuracy above average, 2) coverage above average and 3) length below average
\EndFor
\State $\mathcal{U} \leftarrow \bigcup_{c \in C} \mathcal{I}_c$
\State \Return $\mathcal{U}$
\end{algorithmic}
\label{alg:forex}
\end{algorithm}

\subsubsection{RF+HC}

The RF+HC method can easily be adapted to our budgeted maximum coverage problem. We start by initializing the variable representing the set of selected rules as the empty set, then iteratively add rules. In each iteration, we randomly pick a rule that has not been picked yet and add the rule to the set if it does not violate any budget constraint and increases the coverage of the set. Rules with a high score must be prioritized, so we select rules using a probability distribution proportional to the rule score distribution. Once all rules have been selected and examined, the iterative procedure stops. The result of each run is stored and eventually compared with the results obtained out of a total of $n_\mathrm{trials}$ runs. Since all solutions obtained satisfy the budget constraints, we simply return as unique solution the set that represents the best explanation, i.e. the rule set with highest coverage. See Algorithm \ref{alg:adapted RF+HC}.

\begin{algorithm}[h]
\caption{Adaptation of RF+HC}
\begin{algorithmic}[1]
\State compute the ruleScore of each rule:
$$ \mathrm{ruleScore} = \frac{cc-ic}{cc+ic} + \frac{cc}{ic+4} + \frac{cc}{rl}$$
\State initialize the best feasible solution found $\mathcal{X}_\mathrm{best}$ and its coverage $f^*(\mathcal{X}_\mathrm{best})$: $\mathcal{X}_\mathrm{best} \leftarrow \{ \}$, $f^*_\mathrm{max} \leftarrow 0$
\For{trial from $1$ to $n_\mathrm{trials}$}
\State initialize $\mathcal{X}$ as the empty set
\State initialize $\mathcal{R}$ as the set of all generated rules
\While{$\mathcal{R}$ is not empty}
\State select randomly a rule $R$ from $\mathcal{R}$ according to the probability distribution $P:\mathcal{R} \rightarrow [0,1]$ defined as $P(R) = \frac{\mathrm{ruleScore}(R)}{\sum_{R' \in \mathcal{R}} \mathrm{ruleScore}(R')}$
\If{adding $R$ to $\mathcal{X}$ increases $f^*(\mathcal{X})$ without violating the complexity and error budget constraints}
\State add $R$ to $\mathcal{X}$
\EndIf
\State remove $R$ from $\mathcal{R}$
\EndWhile
\State if $f^*(\mathcal{X}) > f^*_\mathrm{max}$ then do $f^*_\mathrm{max} \leftarrow f^*(\mathcal{X})$ and $\mathcal{X}_\mathrm{best} \leftarrow \mathcal{X}$
\EndFor
\State \Return $\mathcal{X}_\mathrm{best}, f^*_\mathrm{max}$
\end{algorithmic}
\label{alg:adapted RF+HC}
\end{algorithm}

\subsubsection{IRFRE}

The IRFRE method can also be adapted to our problem. An initial population is created by randomly generating bit strings of length $M$ and filtering the corresponding rule sets that fulfil the budget constraints. A multi-objective genetic algorithm involving three objective functions is then applied. The objectives are the coverage, the complexity and number of classification errors of the rule set. The fitness function used is based on the rank of the individuals with respect to the objectives. For the reproduction mechanism, each pair of individuals selected gives birth to two children after uniform crossover and random bitflip mutation, but only children that satisfy the constraints are added to the population. The $n_\mathrm{max-population}$ fittest individual amongst the individuals are selected for the next generation, where $n_\mathrm{max-population}$ is the maximum population size allowed. The evolutionary process is repeated over $n_\mathrm{generations}$ and the algorithm finally returns the individual with highest coverage. See Algorithm \ref{alg:adapted irfre}.

\begin{algorithm}[h]
\caption{Adaptation of IRFRE}
\begin{algorithmic}[1]
\State generate initial population of random binary strings with length $M$ that fulfil the budget constraints
\For{generation from $1$ to $n_\mathrm{generations}$}
\State calculate for each individual the coverage $f^*(r)$, the complexity $L(r)$ and number of classification errors $E(r)$
\State compute the rank of each individual with respect to the 3 objectives
\State select pairs of individuals randomly based on their rank
\State mate the selected pairs of individuals, each pair gives birth to two children after uniform crossover and random bitflip mutation, add the children to the population if they fulfil the budget constraints
\State select the $n_\mathrm{max-population}$ fittest individuals for the next generation
\EndFor
\State \Return the individual $r$ with highest coverage $f^*(r)$ in the final population
\end{algorithmic}
\label{alg:adapted irfre}
\end{algorithm}

\subsection{Genetic Algorithm}

Solving the problem does not require a multi-objective genetic algorithm and its rank-based fitness function and can be solved using a standard (single-objective) genetic algorithm \cite{goldberg1989genetic} with some slight modifications to account for the constraints. We encode the variable $r$ as a bit string $r_1 r_2 \dotsc r_M$ of length $M$ or \enquote{chromosome}.  The encoding means that if $r_i = 1$, then the $i$-th rule $R_i$ is selected for the set of rules, where the set of rules $\mathcal{R}=\{ R_1, \dotsc, R_M\}$ is generated as described in Section \ref{section-generation}.

The fitness function to maximize is the coverage $$f^*(r)=\sum_{j=1}^N e_j^*(r) - \frac{\epsilon}{2} \sum_{j=1}^N \sum_{i=1}^M \sum_{i'=1}^M \alpha_{i,i',j} \cdot r_i \cdot r_{i'}$$ where $e_j^*(r)=1$ when there is a rule selected which explains $x_j$, and $e_j^*(r)=0$ otherwise. Remark that $e_j^*(r)=\mathrm{sign}(e_{j,\bullet}\T r)$, where
\begin{itemize}
    \item $\mathrm{sign}$ is the sign function defined as $\mathrm{sign}(z) = 1$ if $z>0$ and $\mathrm{sign}(z) = 0$ otherwise,
    \item $e_{j, \bullet}$ is the vector in $\mathbb{R}^M$ with $i$-th component $e_{j,i}=1$ if $x_j$ is explained by $R_i$ and $e_{j,i}=0$ otherwise.
\end{itemize}
Observe also that the triple sum $$\sum_{j=1}^N \sum_{i=1}^M \sum_{i'=1}^M \alpha_{i,i',j} \cdot r_i \cdot r_{i'}=r\T \Q^{\alpha} r$$ is a quadratic function, where $\Q^{\alpha}$ is the $M \times M$ square matrix with integer coefficients $\ \Q^{\alpha}_{i,i'} = \sum_{j=1}^N \alpha_{i, i', j}$. Introducing the binary $M \times M$ matrix $\E$ with coefficients $e_{j,i}$ equal to $1$ if and only if $x_j$ is explained by $R_i$, we can rewrite the fitness function in matrix notation as
\begin{equation}\label{fitness_implementation}
    f^*(r) = 1\T \mathrm{sign}(\E) - \frac{\epsilon}{2} \cdot r\T \Q^{\alpha} r
\end{equation}
where the sign function is applied component-wise. The matrix notation allows an efficient implementation using a matrix calculus library such as Python's Numpy.

The crossover mechanism used can be any of the usual ones for genetic algorithms \cite{goldberg1989genetic}, namely:
\begin{itemize}
    \item 1-point crossover: an integer position on both parents' chromosomes is picked randomly and designated a `crossover point'. Bits to the right of that point are swapped between the two parent chromosomes.
    \item 2-point crossover: two crossover points are picked randomly from the parent chromosomes. The bits in between the two points are swapped between the parent chromosomes.
    \item Uniform crossover: each bit is chosen from either parent with equal probability $0.5$.
\end{itemize}
The crossover and mutation operations take place classically as well.

The standard genetic algorithm would not account for the budget constraints. So, at every step of the genetic algorithm, we have to make sure that all individuals satisfy them. This implies removing from the initial population all the individuals that violate a constraint. If after crossover the resulting chromosome does not satisfy a budget constraint, then the operation is restarted until that is the case. This repetition is also necessary when applying random mutations to the chromosomes obtained by crossover. The genetic algorithm used is shown in Algorithm \ref{alg:genetic-algorithm}.

\begin{algorithm}
\caption{Standard Genetic algorithm}
\begin{algorithmic}[1]
\State create a random initial population that fulfils the constraints
\For{$t$ from $1$ to $n_\mathrm{generations}$}
\State evaluate the fitness $f^*(r)$ of each individual $r$
\State let $n$ be the current size of the population, draw $n/2$ pairs of distinct individuals with probability $p(r)=\frac{f^*(r)}{\sum_{r'} f^*(r')}$ proportional to the fitness of each individual (an individual may belong to several pairs)
\State generate 2 children per pair by crossover (repeat until each child satisfies the constraints)
\State apply a mutation to the children (repeat until each child satisfies the constraints)
\State evaluate the fitness of the children
\State add all the children to the current population
\State select only the fittest $n_\mathrm{max-population}$ individuals  
\EndFor
\State \Return the individual with the highest fitness
\end{algorithmic}
\label{alg:genetic-algorithm}
\end{algorithm}

\subsection{Inner Genetic Algorithms}

The genetic algorithm Algorithm \ref{alg:genetic-algorithm} requires a) to generate a random population that fulfils the constraints and b) to generate mutations that keep the constraints satisfied. We will see that these two mathematical problems are in fact NP-hard and will thus introduce two additional genetic algorithms to solve them and nest them inside our outer genetic algorithm. In the overall algorithm, we will first pre-compute the initial population and set of mutations using the inner genetic algorithms. Moreover, the mutations will be applied sequentially on the chromosomes resulting from crossover instead of just once. Finally, the genetic algorithm will be equipped with an early stopping policy. The Algorithm \ref{alg:genetic-algorithm} modified in this way will be referred to as \emph{outer} genetic algorithm.

\subsubsection{Computation of the Initial Population}

For the initial population, we need to find chromosomes that fulfil the two budget constraints. This is mathematically equivalent to finding binary vectors $x$ of dimension $M$ that are solutions of the integer system of two linear inequalities 
$$\A x \leq b~,$$
where $\A$ is the integer matrix of size $2 \times M$
$$\A=\begin{bmatrix}
L(R_1) \dotsc L(R_M) \\
E(R_1) \dotsc E(R_M)
\end{bmatrix}$$
and $b=(B_c, B_e)\T$ is the 2-dimensional budget vector. To obtain solutions, we pick several integer vectors $\beta$ such that $0 \leq \beta \leq b$ and for each $\beta$, we search for exact as well as approximate solutions of $\A x = \beta$. All exact solutions, together with the approximate solutions to $\A x = \beta$ that also satisfy $\A x \leq b$ are then kept as solution of the system and used to form the initial population of the standard genetic algorithm. 

Exact and approximate solutions to $\A x=\beta$ can be obtained by minimizing the squared norm of the difference between $\A x$ and $\beta$
$$\lVert \A x-\beta \rVert^2 = (\A x-\beta)\T(\A x-\beta) = x\T (\A\T \A) x - 2\beta\T \A x + \mathrm{cst}~,$$
where $\mathrm{cst}$ is a constant term independent of $x$. 
Since $x$ is a binary vector, this is equivalent to the Quadratic Unconstrained Binary Optimization (QUBO) problem 
$$\min x\T \Q_B x~, x \in \{0,1\}^{M}$$
where $\Q_B$ is the square integer matrix $\Q_B = \A\T \A - 2 \mathrm{diag} [\beta\T \A]$ of size $M \times M$. A sample of $n$ solutions to the QUBOs can be sampled using a simple genetic algorithm as shown in Algorithm \ref{alg:qubo-genetic-algorithm}. The matrices $\Q_B$ of these QUBOs are of size $M \times M$, where $M$ can in practice be quite large. This issue can be easily circumvented by generating small random subsets of $m \leq M$ dimensions and solving the QUBO in the sub-spaces only. The solutions obtained can be transformed into bit-strings of length $M$ by adding zeros in the positions of the remaining dimensions. The union of all solutions obtained in this way constitutes a sample of solutions of the original QUBO. Our first inner algorithm for generating an initial population is shown in Algorithm \ref{alg:integer-inequality_system}.

\begin{algorithm}
\caption{Solves $\min x\T \Q x~, x \in \{0,1\}^{M}$ by genetic algorithm}
\begin{algorithmic}[1]
\State create $n$ random bit-strings $x$ of length $M$
\For{$t$ from $1$ to $n_\mathrm{generations}$}
\State evaluate the fitness $f(x)=x\T \Q x$ of each individual $x$
\State draw $n/2$ pairs of distinct individuals with probability $p(x)=\frac{f(x)}{\sum_{x'} f(x')}$ proportional to the fitness of each individual (an individual may belong to several pairs)
\State generate 2 children per pair by crossover
\State apply a random mutation to the children
\State evaluate the fitness of the children
\State add all the children to the current population
\State select only the fittest $n$ individuals  
\EndFor
\State \Return the $n$ individuals of the last generation
\end{algorithmic}
\label{alg:qubo-genetic-algorithm}
\end{algorithm}

\begin{algorithm}
\caption{Solves $\A x \leq b$ for $x \in \{0,1\}^M$}
\begin{algorithmic}[1]
\State \textbf{INPUTS}: $\A \in \mathbb{N}^{M \times 2}$, $b \in \mathbb{N}^2$, set $\mathcal{B} \subseteq \{ \beta \in \mathbb{N}^2 \vert 0 \leq \beta \leq b \}$
\State \textbf{OUTPUT}: sample of solutions
\State compute the $M \times M$ matrix $\Q_I = \A\T\A$
\For{$\beta \in \mathcal{B}$}
\State compute the $M \times M$ matrix $\Q_B = \Q_I - 2 \mathrm{diag} [\beta\T A]$
\State sample solutions of the QUBO $\min x\T \Q_B x$, $x \in \{0,1\}^{M}$ by using Algorithm \ref{alg:qubo-genetic-algorithm} in parallel on small sub-spaces of $\{0,1\}^{M}$
\State store in $S$ all sampled $x$ that satisfy $\A x \leq b$
\EndFor
\State remove duplicate solutions from $S$
\State \Return sample of unique solutions
\end{algorithmic}
\label{alg:integer-inequality_system}
\end{algorithm}

\subsubsection{Computation of the Set of Mutations}

We represent a mutation $m$ by a vector in $\{ -1, 0, +1\}^M$ that can suppress, leave unchanged or activate genes. A chromosome $x$ mutates under some mutation $m$ into a new chromosome $x \oplus m$ as $$(x \oplus m)_i \DEF \mathrm{clip}(x_i + m_i, 0, 1)$$ 
where $\mathrm{clip}$ is the clipping function defined as $\mathrm{clip}(z, a, b)=z$ if $a \leq z \leq b$, $\mathrm{clip}(z, a, b)=a$ if $z < a$ and $\mathrm{clip}(z, a, b)=b$ if $z > b$.
Given a solution $x$ of the system $\A x \leq b$, we want to find mutations $m$ such that $x \oplus m$ is still a solution of the system, i.e. $\A (x \oplus m) \leq b$. It is easy to check that $x \oplus m = x + m'$ where $m'$ is the vector in $\{ -1, 0, +1\}^M$ defined as $m'_i = +1$ if $x_i=0$ and $m_i=1$, $m'_i = -1$ if $x_i=1$ and $m_i=-1$, and $m'_i=0$ otherwise. For such a vector $m'$, we have $x \oplus m = x \oplus m' = x + m'$ and consequently 
$$\A (x \oplus m) = \A(x \oplus m') = \A x + \A m'$$
If $m'$ is also such that $\A m' \approx 0$, then $\A(x \oplus m) \approx \A x$, so there is a chance that $x \oplus m'$ still satisfies the inequality system $\A(x \oplus m') \leq b$. In summary, we can obtain good candidates for the mutations by finding elements of the approximate kernel of the integer matrix $\A$ in $\{ -1, 0, +1\}^M$. 

We can solve $\A x \approx 0$ for $x \in \{ -1, 0, +1\}^M$ using again a genetic algorithm. Since each component $x_i \in \{ -1, 0, +1\}$ of the variable $x$ may take three possible values, its encoding requires a minimum of two binary variables. We thus encode each $x_i$ over two binary variables $b_{i,1}$ and $b_{i,2}$ by writing $x_i = -1 + b_{i,1} + 2 b_{i,2}$. It is easy to check that the encoding is such that $(b_{i,1}, b_{i,2})$ is mapped as follows: $(0,0) \mapsto -1, (1,0) \mapsto 0$ and $(0, 1) \mapsto +1$. Our linear encoding can be also written in matrix form:
$$ x = L + \E X = L + (I_M \otimes [1, 2]) X = \begin{bmatrix} 
-1 \\
-1 \\
\vdots \\
-1
\end{bmatrix} + 
\begin{bmatrix}
1 & 2 & 0 & 0 & 0 & \dotsc & 0\\
0 & 0 & 1 & 2 & 0 & \dotsc & 0 \\
\vdots & \\
0 & \dotsc & & & 0 & 1 & 2
\end{bmatrix} 
\begin{bmatrix}
b_{1,1} \\
b_{1,2} \\
b_{2,1} \\
b_{2,2} \\
\vdots \\
b_{M,1} \\
b_{M,2}
\end{bmatrix}
$$
where $L$ is the $M$-dimensional vector $-1 = [-1, \dotsc, -1]$, $X$ is the encoding binary vector of dimension $2M$ obtained by staking all encoding bits $b_{i,1}$ and $b_{i,2}$, $\otimes$ is the Kronecker product and $\E=\mat I_M \otimes [1, 2]$ is the $M \times (2M)$ encoder matrix and $\mat I_M$ the identity matrix of dimension $M$.

Exact and approximate solutions to $\A x=0$ can be obtained by minimizing the squared norm of $\A x$
$$ \lVert \A x \rVert^2 = (\A(L+\E X))\T \A(L+\E X) = X\T \E\T \A\T \A \E X + 2 L\T \A\T \A \E X + \mathrm{cst}~.$$
Since $X$ is a binary vector this is equivalent to the QUBO problem
$$\min X\T \Q_B X, X \in \{0,1\}^{2M}~,$$
where $\Q_B$ is the $2M \times 2M$ square matrix $\Q_B = \E\T Q_I \E + 2 \mathrm{diag}[L\T \Q_I \E]$ and as previously $\Q_I=\A\T\A$. Additional approximate kernel elements can be obtained by closing the sampled approximate kernel under the multiplication by $-1$. Indeed, if $\A x \approx 0$, then $\A(-x) \approx 0$. The approximate kernel of $A$ is computed using Algorithm \ref{alg:approximate-kernel}.

\begin{algorithm}[h]
\caption{Solves $\A x \approx 0$ for $x \in \{ -1, 0, +1\}^M$}
\begin{algorithmic}[1]
\State \textbf{INPUTS}: $\A \in \mathbb{N}^{2 \times M}$
\State \textbf{OUTPUT}: sample of solutions from the approximate kernel of $\A$
\State compute the $M \times M$ matrix $\Q_I = \A\T\A$
\State compute the $M \times (2M)$ encoder matrix: $ \E = \mat I_M \otimes [1, 2]$ 
\State compute the $2M \times 2M$ matrix $\Q_B = \E\T \Q_I \E + 2 \mathrm{diag}[L\T \Q_I \E]$
\State sample solutions of the QUBO $\min X\T \Q_B X$, $X \in \{0,1\}^{2M}$ using Algorithm \ref{alg:qubo-genetic-algorithm}
\State decode optimal and approximate binary vectors $X$ using $x = L + \E X \in \mathbb{Z}^{M}$
\State store in $S$ all such $x$ that satisfy $-1 \leq x \leq 1$
\State close $S$ under multiplication by $-1$
\State remove duplicate solutions from $S$
\State \Return $S$
\end{algorithmic}
\label{alg:approximate-kernel}
\end{algorithm}

\subsubsection{Super-mutations}

In the standard genetic algorithm (Algorithm \ref{alg:genetic-algorithm}), a random mutation is applied to a chromosome $x$ such that the mutated chromosome fulfils the budget constraints. The mutated chromosome $r$ does not necessarily have a higher fitness. Mutations are applied differently in the case of the nested genetic algorithm. Since the mutations $m$ pre-generated using the inner genetic algorithm Algorithm \ref{alg:approximate-kernel} have a higher chance of preserving the feasibility of the solution $r$, we can impose that $m$ increases the fitness, i.e. $f^*(r \oplus m) > f^*(r)$. We may refer to such mutations as \emph{augmenting} mutations. And instead of applying a single mutation after each crossover, we may apply a sequence $m_1, \dotsc, m_n$ of augmenting mutations such that 
$$ f^*(r) < f^*(r \oplus m_1) < f^*(r \oplus m_1 \oplus m_2) < \dotsc < f^*(r \oplus m_1 \oplus \dotsc \oplus m_n)$$
and stop applying mutations whenever the pre-generated mutations cannot augment the fitness value anymore. We introduce the concept of \emph{super-mutation} to denote a maximal sequence of augmenting mutations. Super-mutations can be seen as an augmentation procedure or hill-climbing method applied to each chromosome obtained by crossover.

\subsubsection{Early Stopping Policy}

The classical genetic algorithm is run for a predetermined number of generations. In the nested genetic algorithm, super-mutations are used and cause the genetic \emph{diversity} (population size when redundant chromosomes are removed) to shrink rapidly. Indeed, all chromosomes obtained by crossover are subject to super-mutations, and we observe experimentally that they converge to a smaller number of local maxima (top of the hills). Thus, the genetic diversity decreases quickly to $1$, which means that all individuals become genetically identical to one individual that we may call the \enquote{dominating mutant}. This can be easily detected without comparing the individuals' chromosome, but by checking that their fitness value is the same for all individuals. After this point, the crossover operation results in identical offspring and mutations cannot further improve the fitness value anymore. We thus stop the nested genetic algorithm as soon as i) the minimum and maximum fitness values in the population are equal, ii) the pre-specified number of generations is reached or iii) we find an individual with maximal fitness score of $100\%$. The nested genetic algorithm finally obtained is provided for clarity in Algorithm \ref{alg:nested-genetic-algorithm}. We refer to the nested genetic algorithm as \emph{Quadratic Genetic Algorithm}, abbreviated as \enquote{QGA} due to its double use of Quadratic Unconstrained Binary Optimization (QUBO) models.

\begin{algorithm}
\caption{Quadratic Genetic Algorithm}
\begin{algorithmic}[1]
\State create an initial population that fulfils the constraints using the inner genetic algorithm Algorithm \ref{alg:integer-inequality_system}
\State compute a set of mutations using the inner genetic algorithm Algorithm \ref{alg:approximate-kernel}
\For{$t$ from $1$ to $n_\mathrm{generations}$}
\State evaluate the fitness $f^*(r)$ of each individual $r$
\State let $n$ be the current size of the population, draw $n/2$ pairs of distinct individuals with probability $p(r)=\frac{f^*(r)}{\sum_{r'} f^*(r')}$ proportional to the fitness of each individual (an individual may belong to several pairs)
\State generate 2 children per pair by crossover (repeat until each child satisfies the constraints)
\State apply a mutation to the children (repeat until each child satisfies the constraints)
\State repeat for each child the previous step until there is no mutation that can strictly increase its fitness (super mutation)
\State add all the children to the current population
\State select only the fittest $n_\mathrm{max-population}$ individuals
\State exit the for loop if the minimum and maximum fitness values in the population are equal or if we the maximum fitness score is $100\%$
\EndFor
\State \Return the individual with the highest fitness
\end{algorithmic}
\label{alg:nested-genetic-algorithm}
\end{algorithm}

\section{Experiments}\label{section-experiments}

In this section, we describe how the data sets have been selected, the setting used for generating decision rules, how the budget constraints are fixed, summarise performance results and plot the evolution of the fitness value in the quadratic genetic algorithm.

\subsection{Data Sets}

To evaluate the nested genetic algorithm, we compare its performance with other approaches on public data sets from the UCI Machine Learning repository \cite{Dua:2019}. The data sets used must be classification data sets available in tabular form. Since we want the data sets to be explained using sufficiently simple decision rules, we filter for data sets from the repository with a sufficiently small number of attributes (16). As the number of rules required for an explanation increases with the number of classes, we also limit ourselves to data sets with a small number of classes (8). To keep the computation time reasonable, we picked data sets with less than 1,500 instances. From the remaining data sets available, we chose those that could directly be imported without pre-processing work. The left hand side of Table~\ref{tab:experimental-setting} shows the data sets used for the experiments and their respective number number of instances, attributes and classes.

\begin{table}[h]
\centering
\caption{Data sets and setting used in experiments.}
\label{tab:experimental-setting}
\begin{tabular}{|| p{2.5cm} c c c|| c c c||} 
 \hline
 data set & Instances & Attributes & Classes & Depth & Trees & Rules \\ [0.5ex] 
 \hline\hline
 Iris & 150 & 4 & 3 & 3 & 24 & 64 \\ 
 \hline
 Wine & 178 & 13 & 3 & 5 & 48 & 285 \\
  \hline
 Zoo & 101 & 16 & 7 & 5 & 48 & 236 \\
 \hline
 Breast cancer & 699 & 10 & 2 & 5 & 48 & 310 \\
 \hline
 Banknote & 1372 & 4 & 2 & 5 & 48 & 285 \\
 \hline
 Hayes-Roth & 160 & 5 & 3 & 8 & 84 & 1015 \\
 \hline
 Mammographic mass & 961 & 5 & 2 & 7 & 72 & 1247 \\
 \hline
 Blood transfusion & 748 & 5 & 3 & 8 & 84 & 825 \\
 \hline
 Ecoli & 336 & 7 & 8 & 6 & 60 & 587 \\
 \hline
 Tic-tac-toe & 958 & 9 & 2 & 10 & 108 & 5575 \\
 \hline
\end{tabular}
\end{table}

\subsection{Setting used for Generating Decision Rules}

For each data set, we train several decision trees using the CART algorithm implemented in Python's scikit-learn. The hyper-parameter combinations used for the CART algorithm are those that have been exposed in Table \ref{tab:hyper-parameters-table}. The maximum depth for the trees trained ranges from 1 to the number indicated in the column \enquote{Depth} of Table \ref{tab:experimental-setting}. This number was adjusted manually depending on the intrinsic complexity of each data set. For each hyper-parameter combination, one tree is trained and the total number of decision trees trained for each data set is shown is the column \enquote{Trees}. All trees are transformed into decision rules, which are simplified whenever possible by merging conditions on the same features. Decision rules that are redundant in the sense of yielding the same predictions on the data set were removed. The final number of decision rules obtained is indicated in the last column \enquote{Rules} and is comprised between $M=64$ and $M=5575$.

\subsection{Choice of the Complexity and Error Budgets}

For each data set, we compute the complexity of the set of decision rules obtained from each decision tree. We call \emph{best tree}, the tree that achieves the minimum complexity after simplification of the rules. For this set of decision rules, the coverage score is guaranteed to be $100\%$. Let then $c^*$ and $e^*$ be the complexity and number of classification errors of the rule set obtained with the best tree. To make the problem challenging, we then set the complexity budget $B_c$ strictly below $c^*$ and the error budget below $e^*$: $B_c < c^*$ and $B_e \leq e^*$. The budgets chosen for each data set are indicated after the slash character in the columns \enquote{Complexity} and \enquote{Errors} of Table \ref{tab:experimental-results}.

\subsection{Performance Results}

\begin{figure}[htbp]
\centering
\subfloat[]{\includegraphics[width=0.5\textwidth, height=3cm]{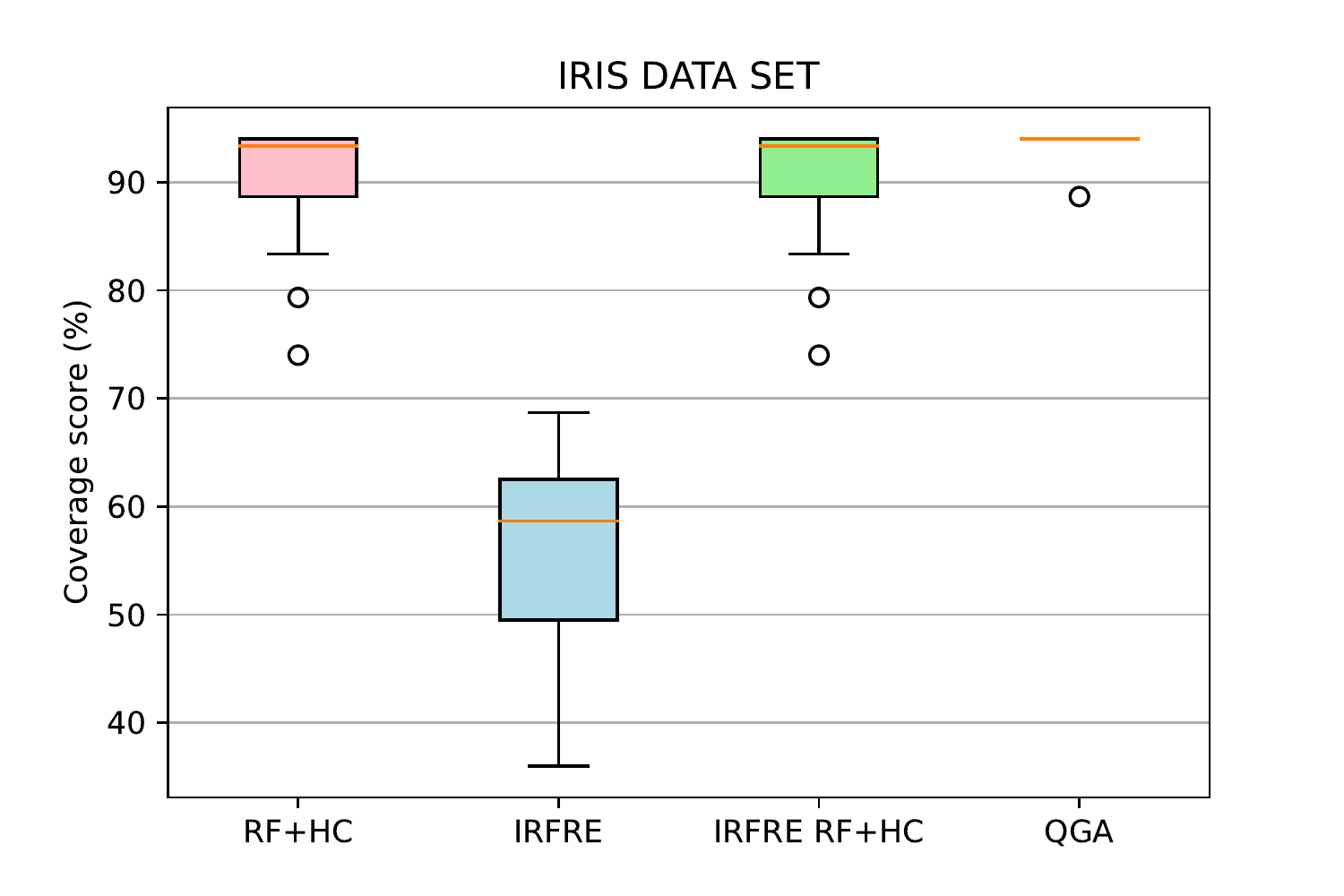}}
  \subfloat[]{\includegraphics[width=0.5\textwidth, height=3cm]{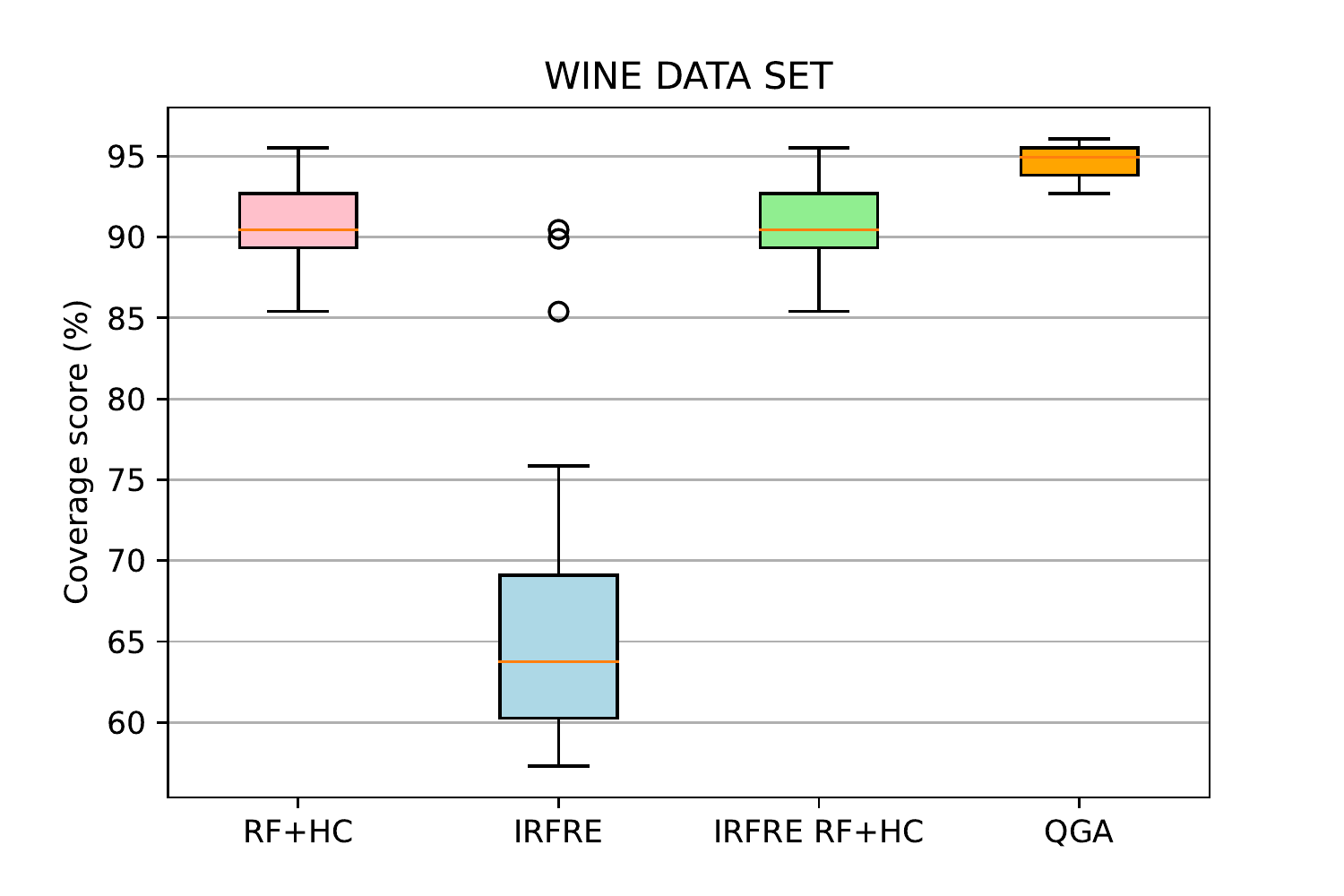}}

  \subfloat[]{\includegraphics[width=0.5\textwidth, height=3cm]{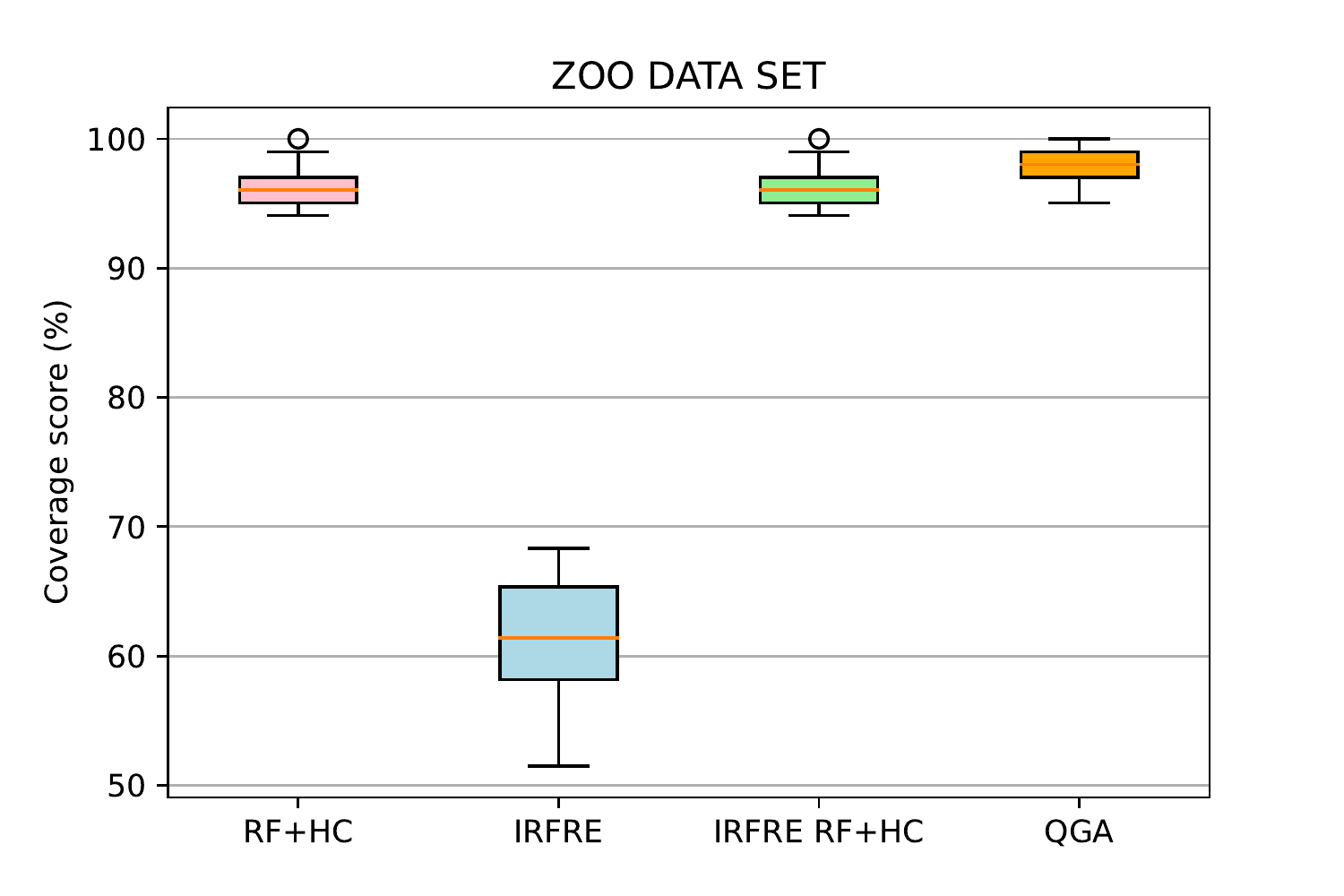}}
  \subfloat[]{\includegraphics[width=0.5\textwidth, height=3cm]{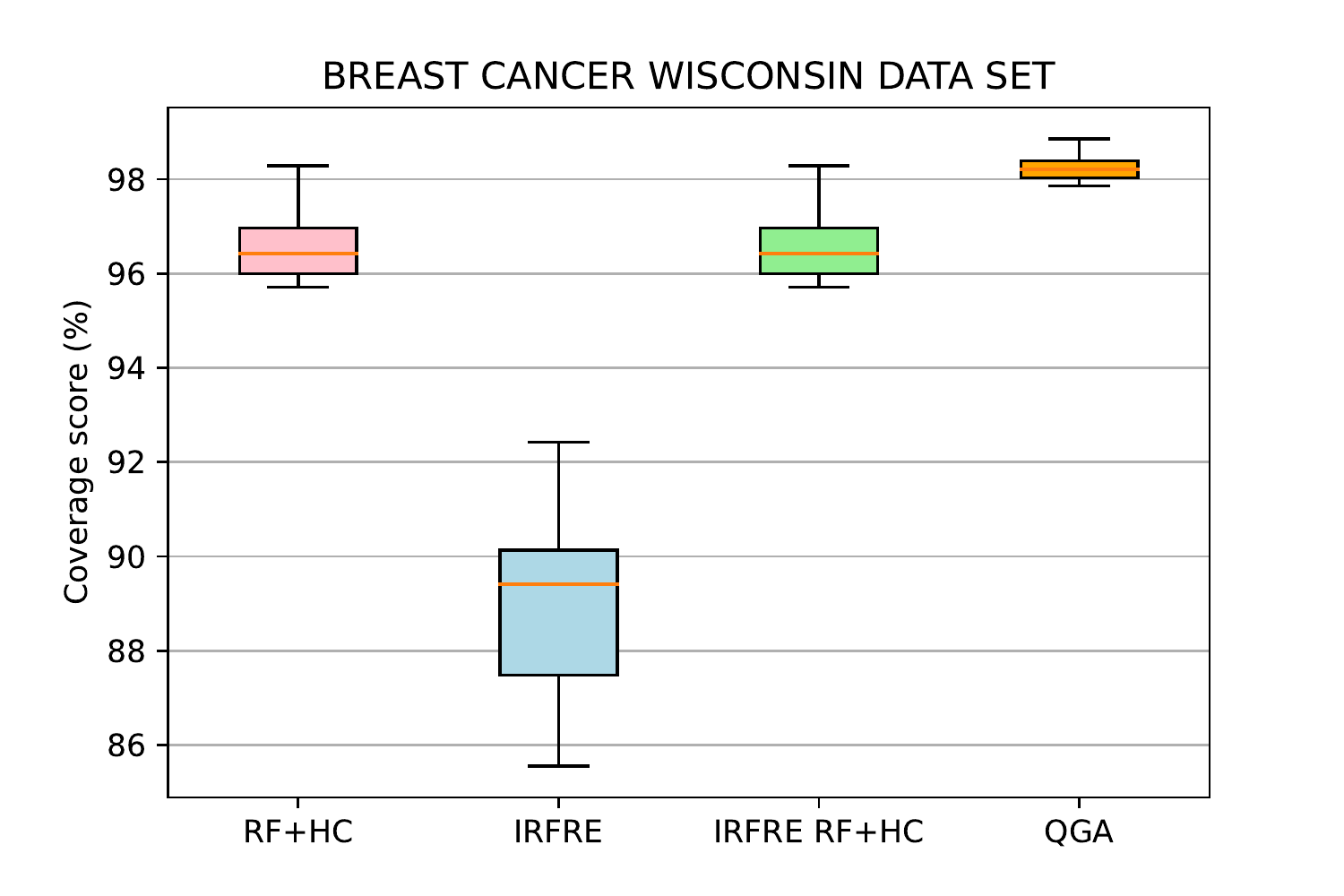}}

  \subfloat[]{\includegraphics[width=0.5\textwidth, height=3cm]{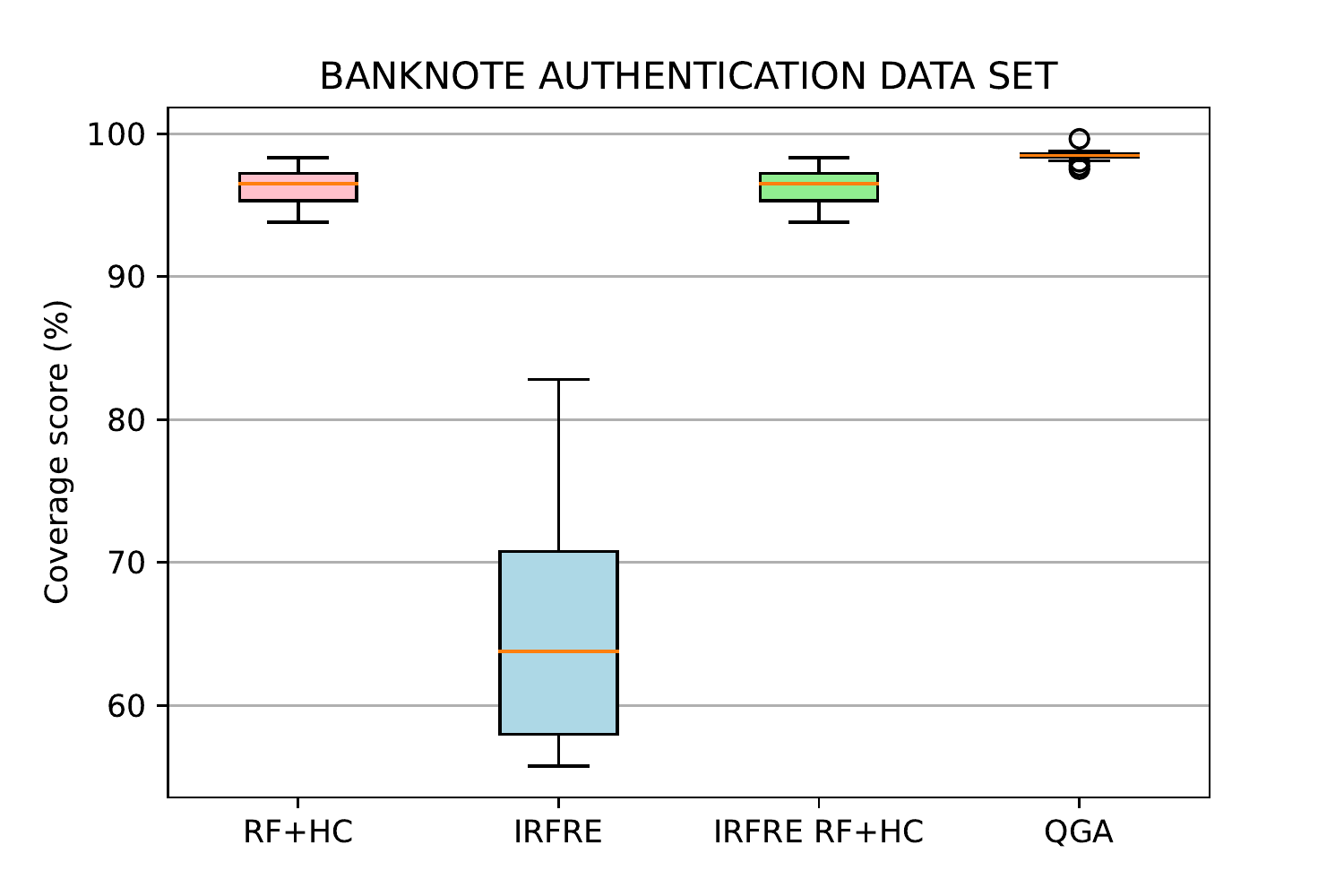}}
  \subfloat[]{\includegraphics[width=0.5\textwidth, height=3cm]{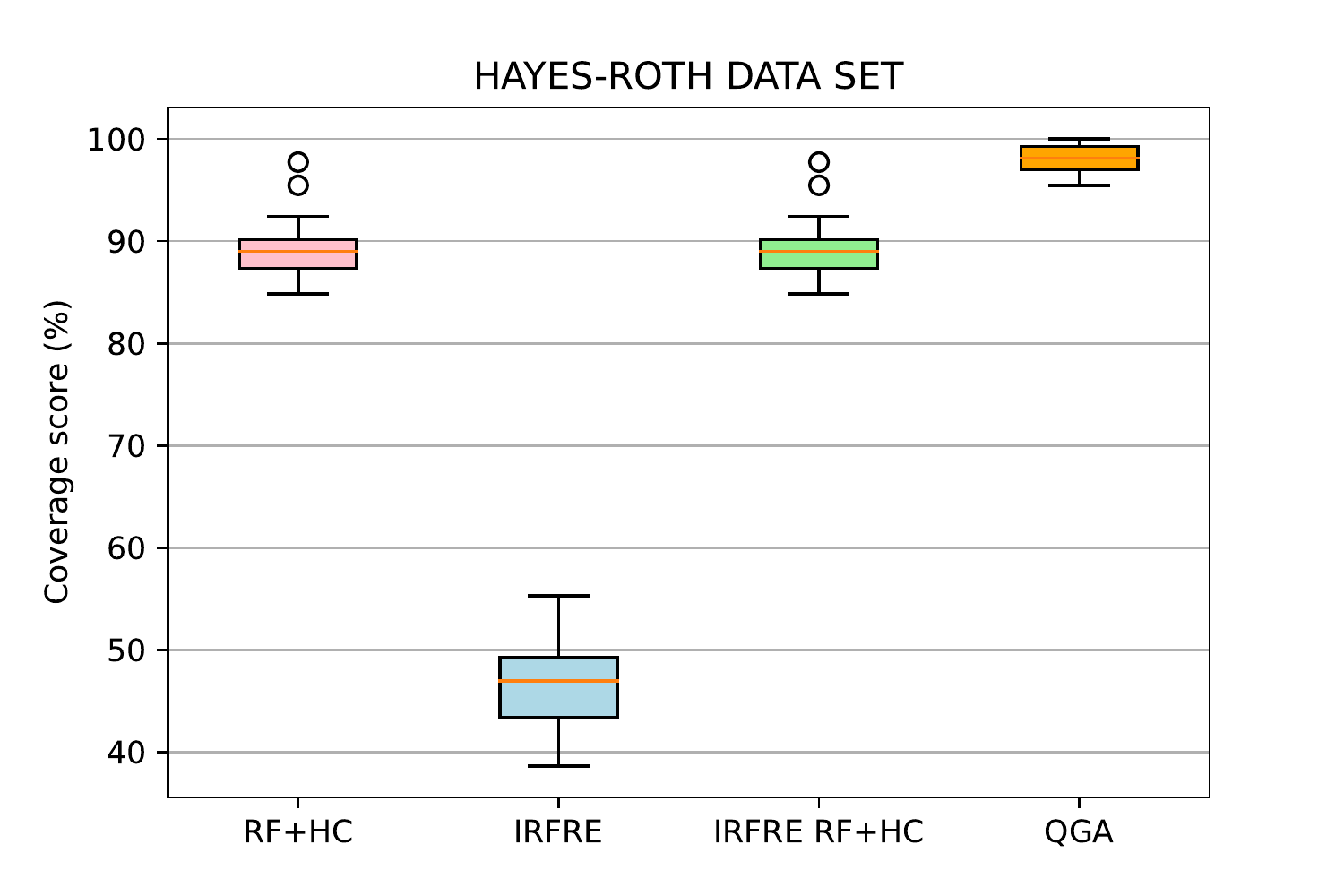}}
  
  \subfloat[]{\includegraphics[width=0.5\textwidth, height=3cm]{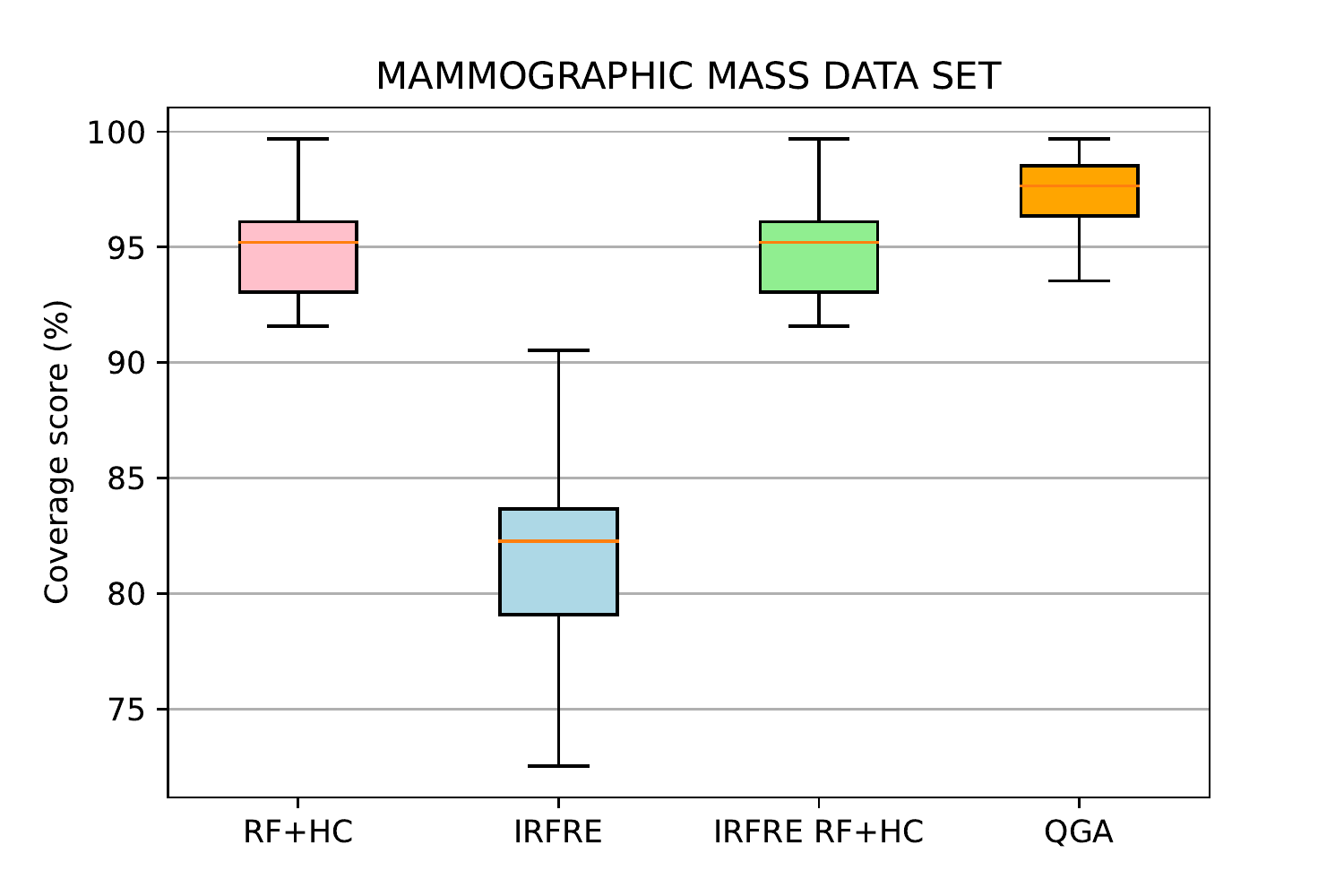}}
  \subfloat[]{\includegraphics[width=0.5\textwidth, height=3cm]{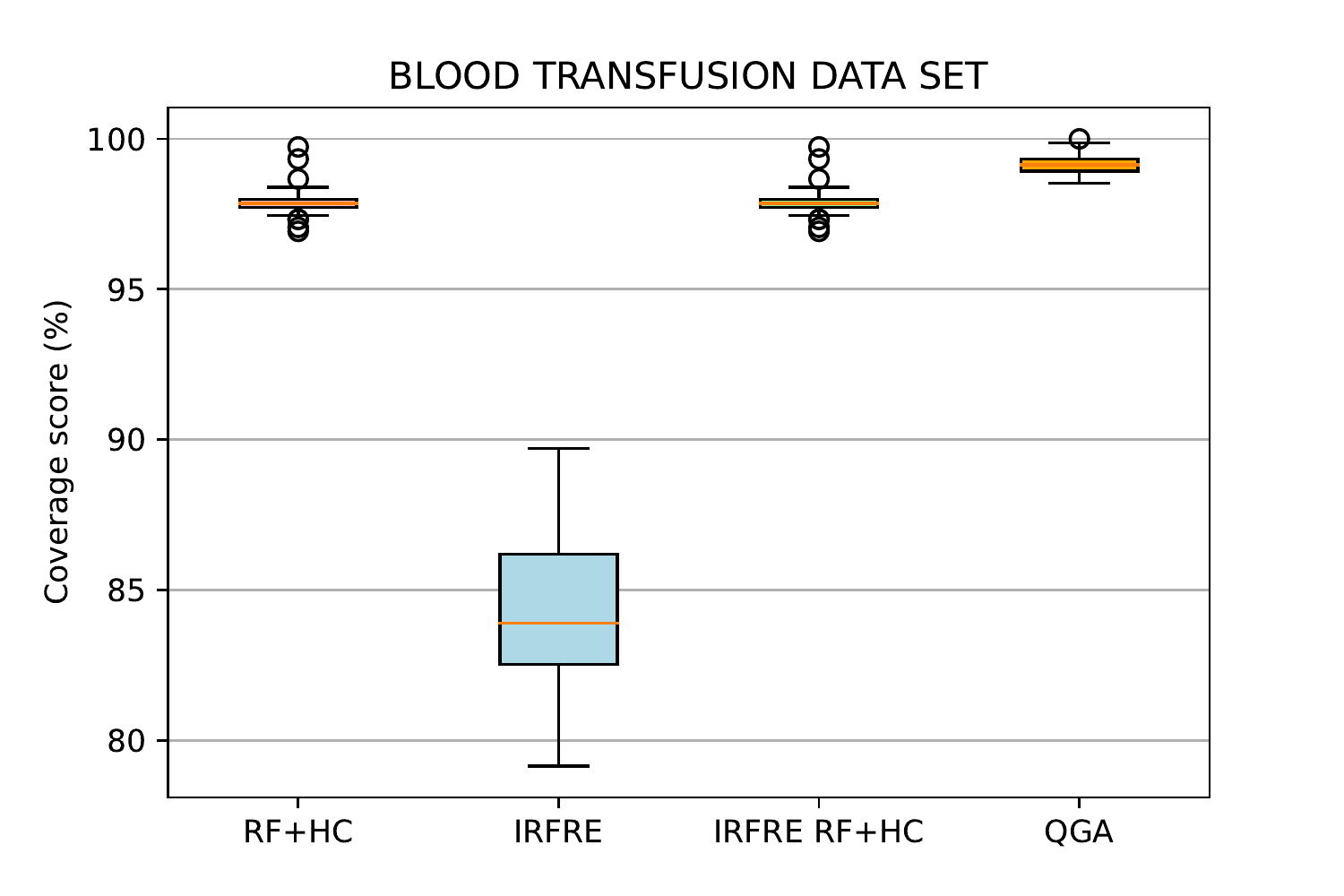}}
  
  \subfloat[]{\includegraphics[width=0.5\textwidth, height=3cm]{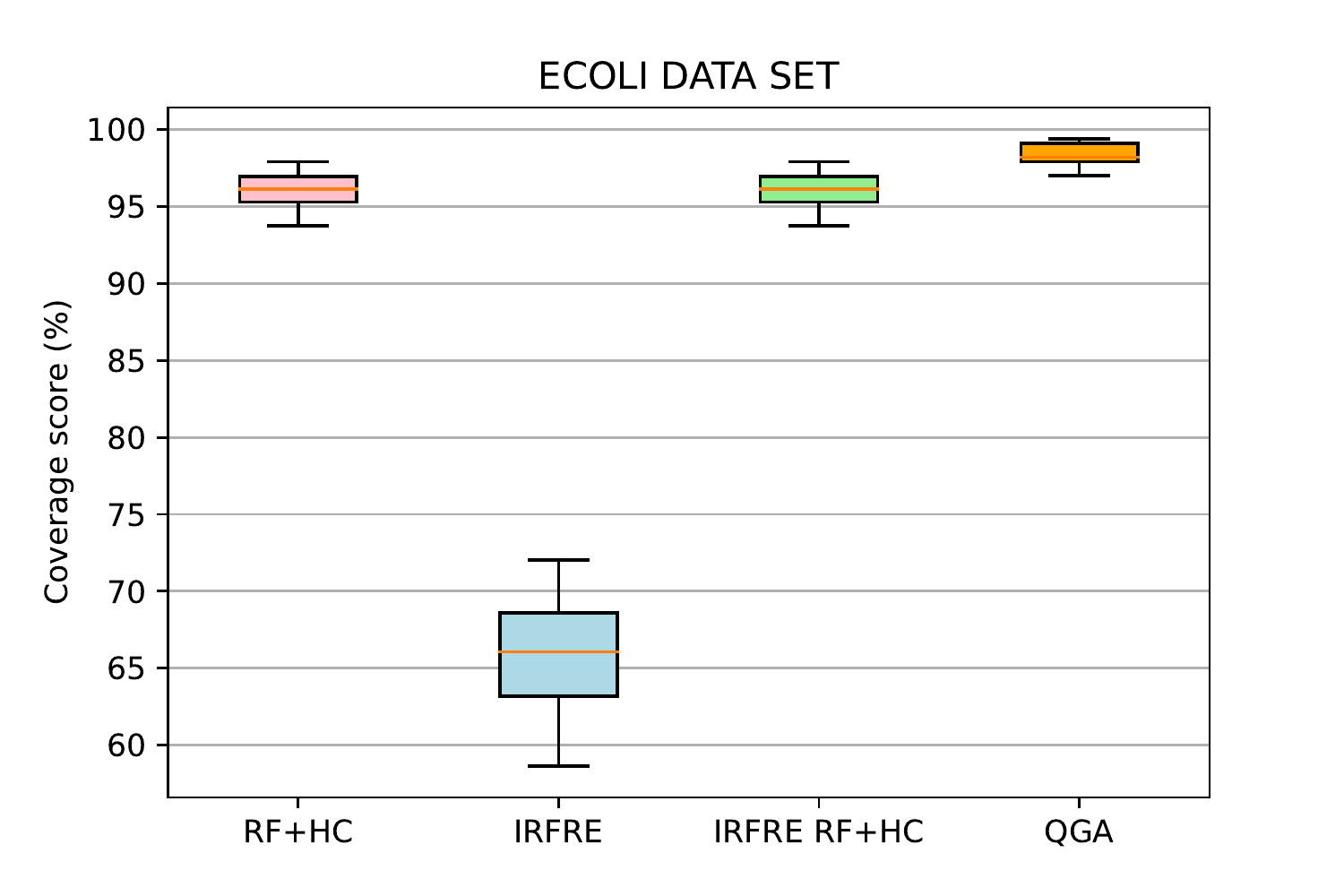}}
  \subfloat[]{\includegraphics[width=0.5\textwidth, height=3cm]{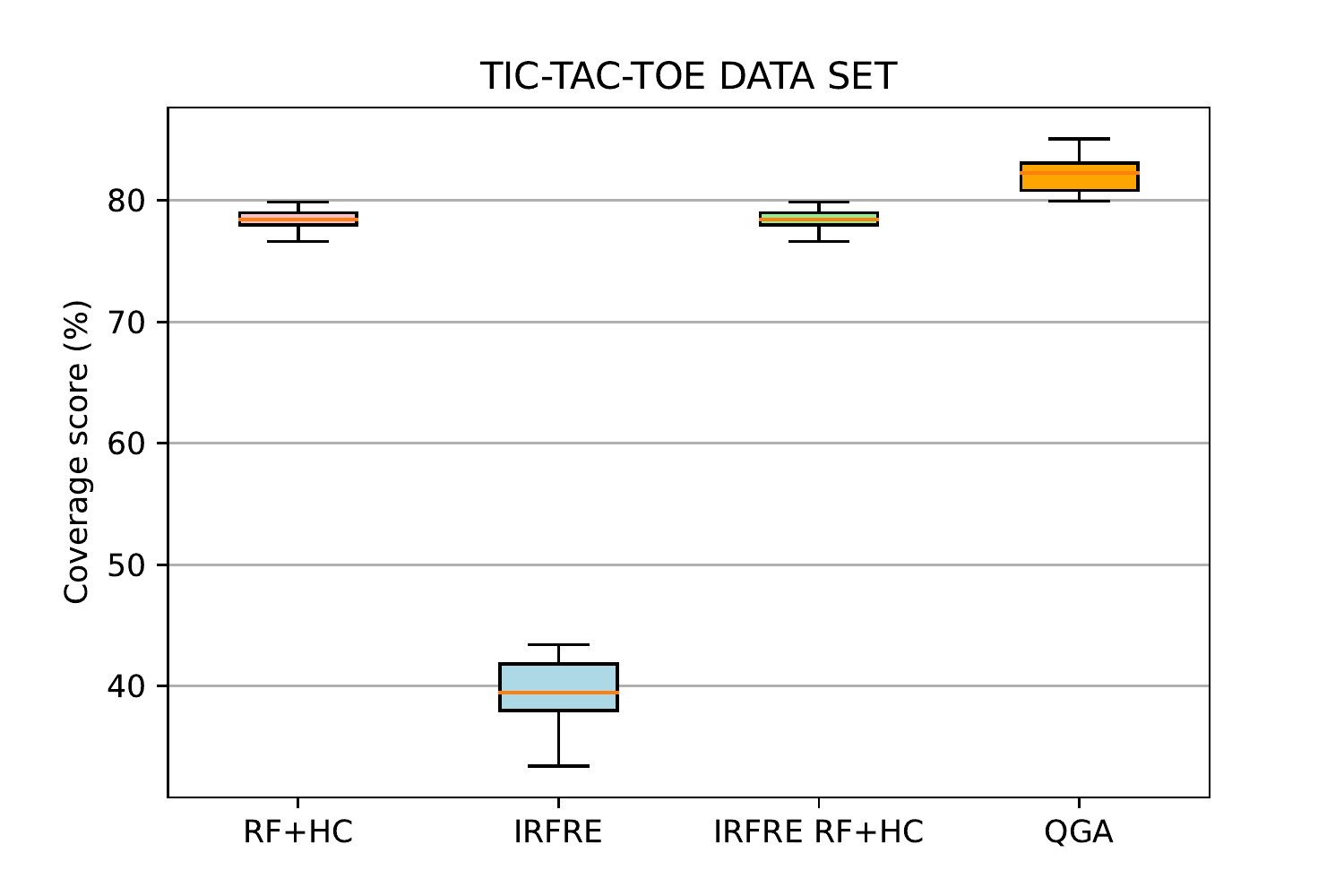}}
\caption{Performance of the selection algorithms.From left to right: RF+HC (pink), IRFRE (blue), IRFRE RF+HC (green) and QGA (orange).}
    \label{fig:barchart}
\end{figure}

For each data set, we evaluate the performance of four algorithms, namely the adaptations of RF+HC and IRFRE, an improvement of the adapted version of IRFRE called IRFRE RF+HC whereby all the results of RF+HC are added to the initial population of the multi-objective genetic algorithm and the nested genetic algorithm (QGA). For each run of these algorithms, we measure the complexity and number of classification errors of the rule set, the computation time and the coverage score. Since the outcomes are all random, we average the measurements over 30 runs. The results are shown in Table \ref{tab:experimental-results} and Figure \ref{fig:barchart} and can be interpreted as follows.

\begin{table}[htbp]
\centering
\caption{Averaged results obtained over 30 runs.}
\label{tab:experimental-results}
\begin{tiny}
\begin{tabular}{||p{2.2cm} |c |c |c |c |c||}
  \hline
Data set & Algorithm & Complexity & Errors & Time & Coverage \% \\ [0.5ex] 
 \hline \hline
Iris & RF+HC & 4.0 / 4 & 3.0 / 5 & 3.3s & 91.0($\pm$4.7) \\ 
& IRFRE & 3.6 / 4 & 2.2 / 5 & 9.2s & 56.4($\pm$9.1) \\ 
& IRFRE RF+HC & 4.0 / 4 & 3.0 / 5 & 18.0s & 91.0($\pm$4.7) \\ 
& \textbf{QGA} & 4.0 / 4 & 2.1 / 5 & 72.5s & \textbf{93.8}($\pm$0.9) \\ 
\hline
Wine & RF+HC & 8.0 / 8 & 7.5 / 9 & 6.0s & 90.8($\pm$2.5) \\ 
& IRFRE & 6.9 / 8 & 5.4 / 9 & 12.4s & 66.7($\pm$8.9) \\ 
& IRFRE RF+HC & 8.0 / 8 & 7.5 / 9 & 22.2s & 90.8($\pm$2.5) \\ 
& \textbf{QGA} & 7.7 / 8 & 8.1 / 9 & 90.8s & \textbf{94.9}($\pm$1.0) \\ 
\hline
Zoo & RF+HC & 17.3 / 18 & 7.9 / 8 & 2.1s & 96.2($\pm$1.5) \\ 
& IRFRE & 7.8 / 18 & 5.2 / 8 & 9.5s & 61.5($\pm$4.3) \\ 
& IRFRE RF+HC & 17.3 / 18 & 7.9 / 8 & 19.1s & 96.2($\pm$1.5) \\ 
& \textbf{QGA} & 17.1 / 18 & 7.8 / 8 & 82.6s & \textbf{98.2}($\pm$1.3) \\ 
\hline 
Breast cancer & RF+HC & 19.5 / 20 & 29.7 / 30 & 18.6s & 96.6($\pm$0.8) \\ 
& IRFRE & 12.3 / 20 & 23.7 / 30 & 46.0s & 89.0($\pm$1.9) \\ 
& IRFRE RF+HC & 19.5 / 20  & 29.7 / 30 & 83.7s & 96.6($\pm$0.8) \\ 
& \textbf{QGA} & 18.5 / 20 & 28.6 / 30 & 117.1s & \textbf{98.3}($\pm$0.3) \\ 
\hline
Banknote & RF+HC & 22.6 / 23 & 34.1 / 35 & 24.3s & 96.3($\pm$1.2) \\ 
& IRFRE & 11.0 / 23 & 21.5 / 35 & 51.4s & 64.9($\pm$6.9) \\ 
& IRFRE RF+HC & 22.6 / 23 & 34.1 / 35 & 90.9s & 96.3($\pm$1.2) \\ 
& \textbf{QGA} & 22.6 / 23 & 31.7 / 35 & 111.9s & \textbf{98.5}($\pm$0.3) \\ 
\hline
Hayes-Roth & RF+HC & 73.4 / 74 & 13.9 / 14 & 15.7s & 89.2($\pm$2.7) \\ 
& IRFRE & 39.9 / 74 & 12.5 / 14 & 23.8s & 46.7($\pm$4.4) \\ 
& IRFRE RF+HC & 73.4 / 74 & 13.9 / 14 & 36.8s & 89.2($\pm$2.7) \\ 
& \textbf{QGA} & 69.5 / 74 & 13.8 / 14 & 136.2s & \textbf{98.0}($\pm$1.4) \\ 
\hline
Mammographic mass & RF+HC & 89.2 / 90 & 150.0 / 150 & 347.3s & 94.9($\pm$2.0) \\ 
& IRFRE & 46.6 / 90 & 136.1 / 150 & 453.7s & 81.9($\pm$4.2) \\ 
& IRFRE RF+HC & 89.2 / 90 & 150.0 / 150 & 601.5s & 94.9($\pm$2.0) \\ 
& \textbf{QGA} & 84.6 / 90 & 148.7 / 150 & 690.9s & \textbf{97.3}($\pm$1.5) \\ 
\hline
Blood transfusion & RF+HC & 96.4 / 100 & 150.0 / 150 & 231.6s & 97.9($\pm$0.6) \\ 
& IRFRE & 40.1 / 100 & 141.4 / 150 & 340.4s & 84.4($\pm$2.8) \\ 
& IRFRE RF+HC & 96.4 / 100 & 150.0 / 150 & 480.3s & 97.9($\pm$0.6) \\ 
& \textbf{QGA (simulator)} & 96.9 / 100 & 149.4 / 150 & 298.0s & \textbf{99.1}($\pm$0.3) \\ 
\hline
Ecoli & RF+HC & 107.5 / 113 & 31.0 / 31 & 26.4s & 96.1($\pm$1.1) \\ 
& IRFRE & 24.4 / 113 & 24.8 / 31 & 42.3s & 66.0($\pm$3.6) \\ 
& IRFRE RF+HC & 107.5 / 113 & 31.0 / 31 & 67.3s & 96.1($\pm$1.1) \\ 
& \textbf{QGA} & 106.3 / 113 & 30.4 / 31& 183.2s & \textbf{98.3}($\pm$0.7) \\ 
\hline
Tic-tac-toe & RF+HC & 96.3 / 150 & 70.0 / 70 & 29.0s & 78.4($\pm$0.8) \\ 
& IRFRE & 14.4 / 150 & 62.8 / 70 & 48.9s & 39.5($\pm$2.5) \\ 
& IRFRE RF+HC & 96.3 / 150 & 70.0 / 70 & 80.8s & 78.4($\pm$0.8) \\ 
& \textbf{QGA} & 112.9 / 150 & 69.5 / 70 & 113.4s & \textbf{82.3}($\pm$1.4) \\ 
\hline
\end{tabular}
\end{tiny}
\end{table}

The RF+HC method is fast and achieves in all cases a high coverage score. The IRFRE method yields feasible solutions but low coverage scores. The IRFRE RF+HC method is significantly better than IRFRE in terms of coverage score, but no better than RF+HC. This happens because the evolutionary process of the multi-objective genetic algorithm does not work as expected. Indeed, the chromosomes resulting from crossover and random mutations are rarely feasible solutions, and when they are, they do not offer any improvement of the fitness value. Thus, both in IRFRE and IRFRE RF+HC, the population remains unchanged after each generation. This issue is resolved by our nested genetic algorithm, which is able to improve on the RF+HC solutions injected in the initial population and allows to reach averaged coverage scores comprised between $90$ and $98.6\%$. The inner genetic algorithms offer better sampling of the solutions of the inequality system $\A x \leq b$ and approximate kernel $\A x \approx 0$, resulting in higher coverage scores whilst reducing the computation time.

\subsection{Fitness Evolution in the QGA}

\begin{figure}[h]
  \centering
  \subfloat[Iris]{\includegraphics[width=0.5\textwidth, height=3cm]{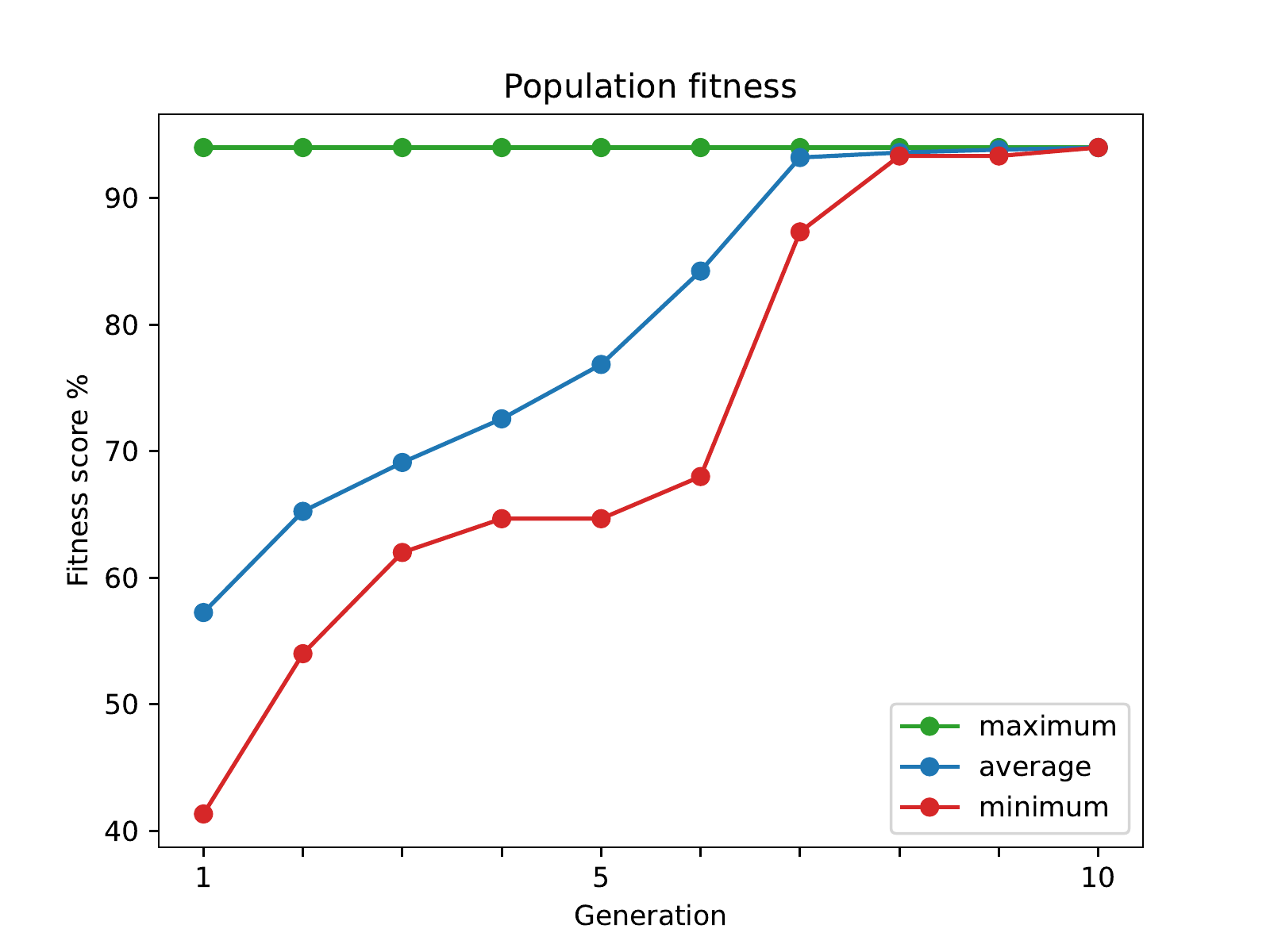}}
  \subfloat[Wine]{\includegraphics[width=0.5\textwidth, height=3cm]{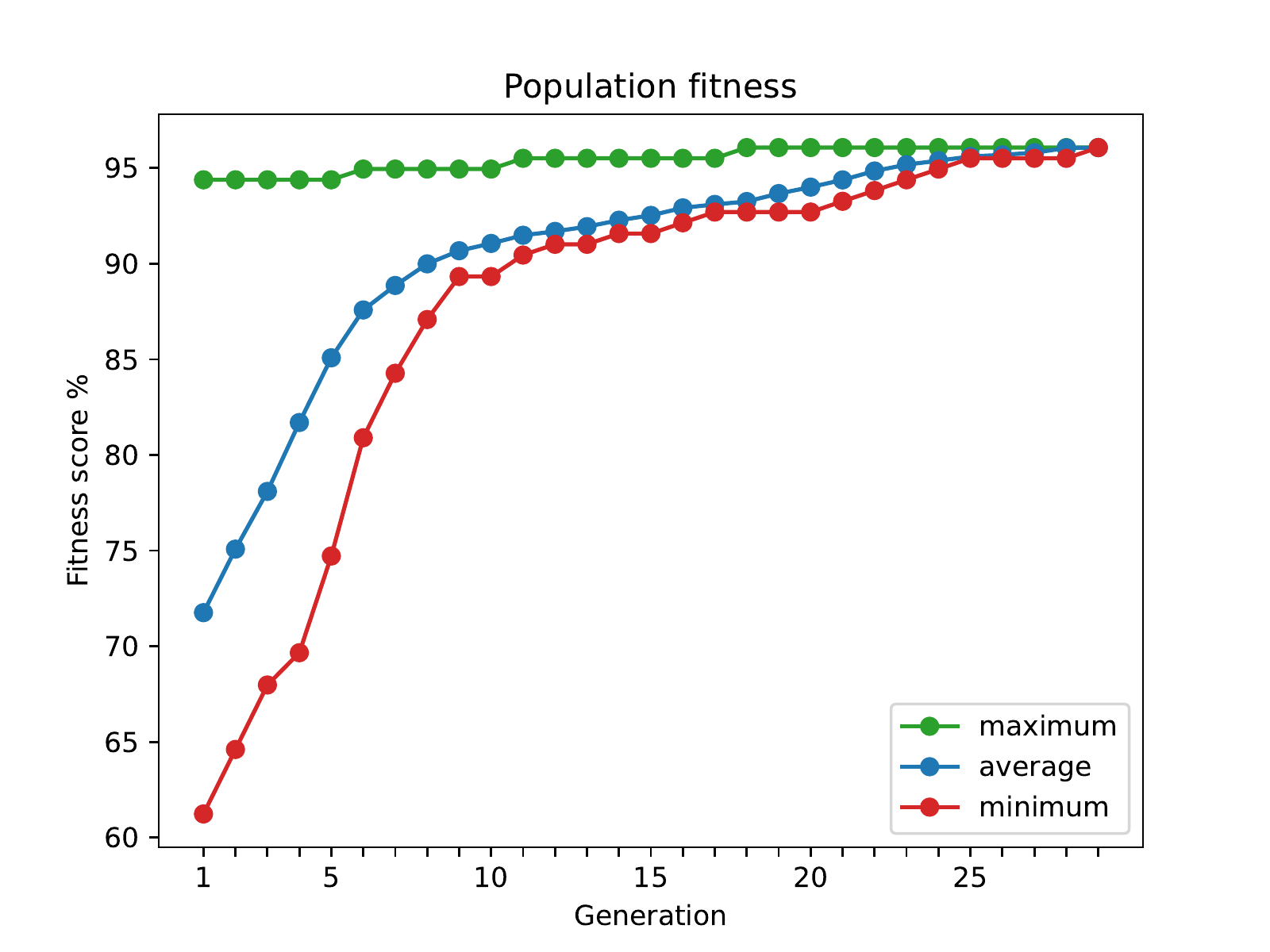}}

  \subfloat[Zoo]{\includegraphics[width=0.5\textwidth, height=3cm]{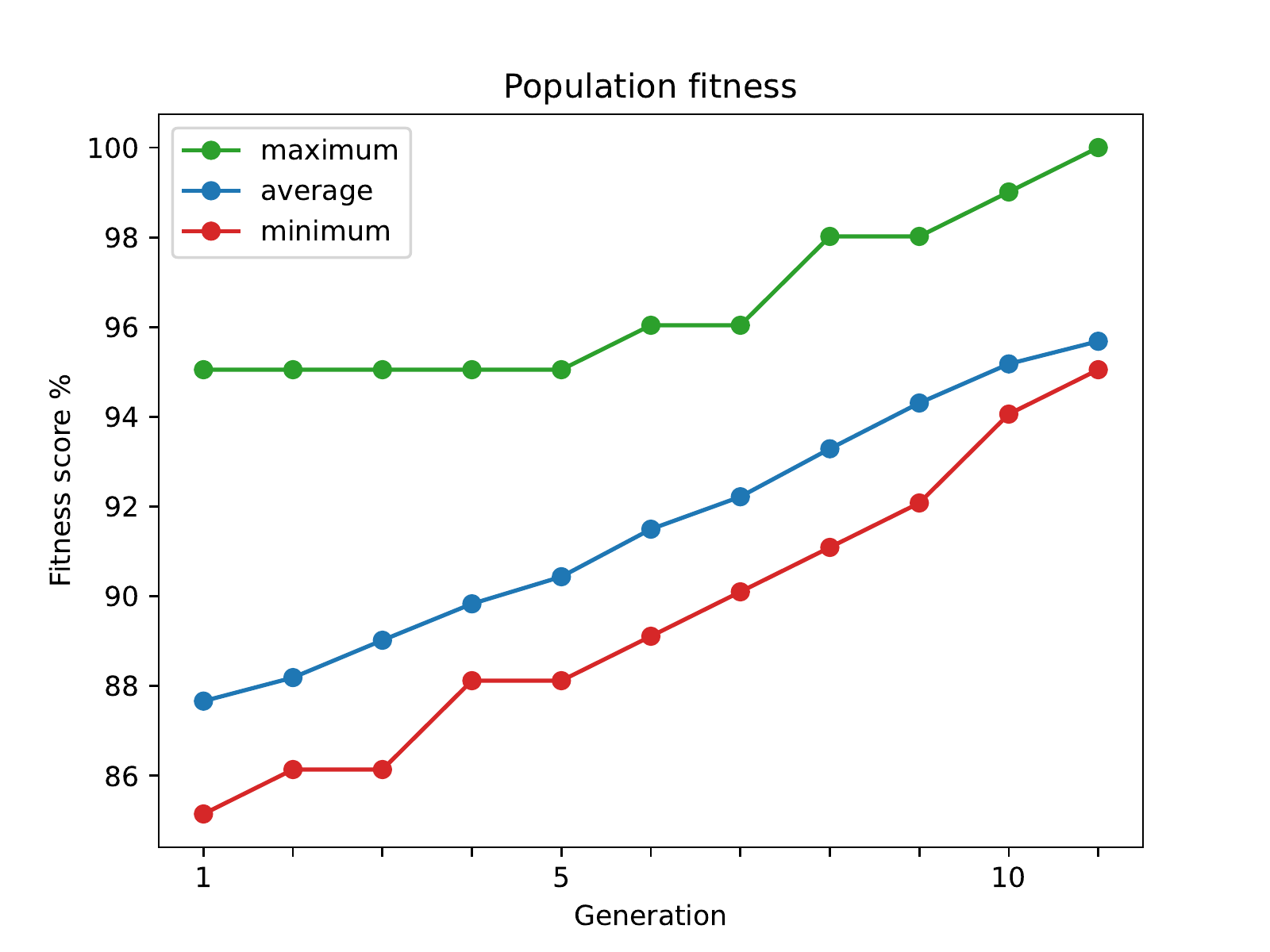}}
  \subfloat[Breast Cancer Wisconsin]{\includegraphics[width=0.5\textwidth, height=3cm]{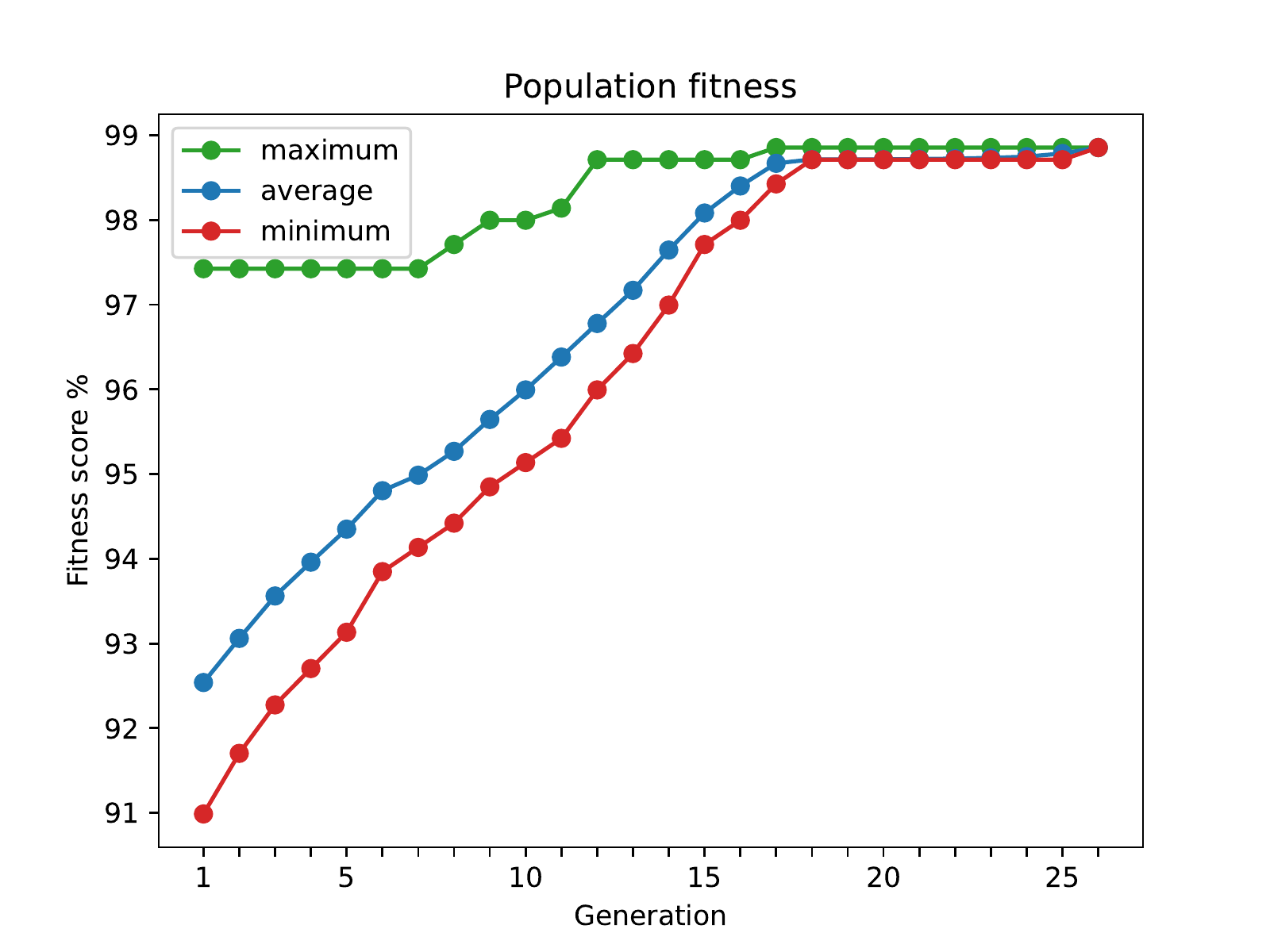}}

  \subfloat[Banknote Authentication]{\includegraphics[width=0.5\textwidth, height=3cm]{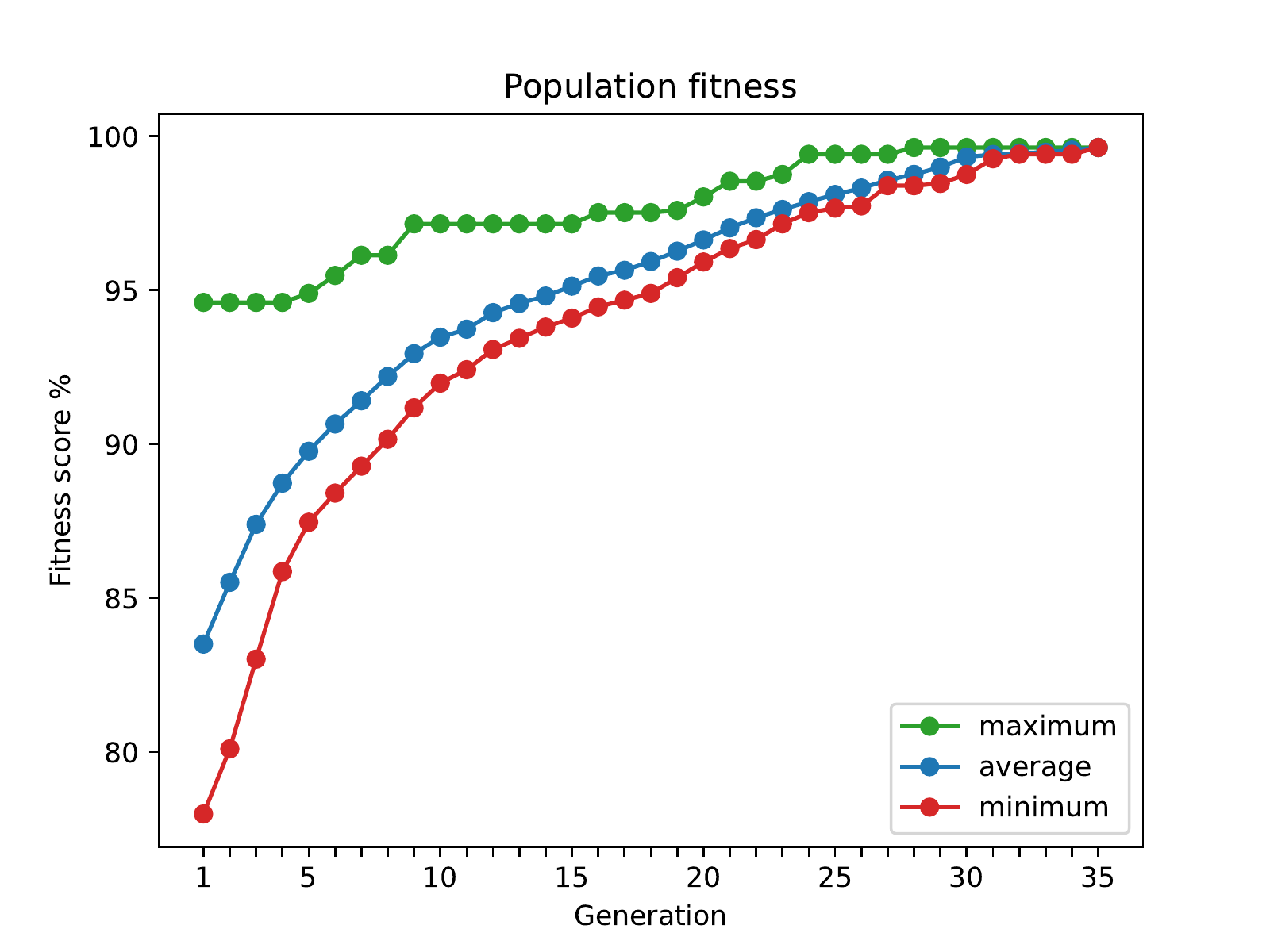}}
  \subfloat[Hayes-Roth]{\includegraphics[width=0.5\textwidth, height=3cm]{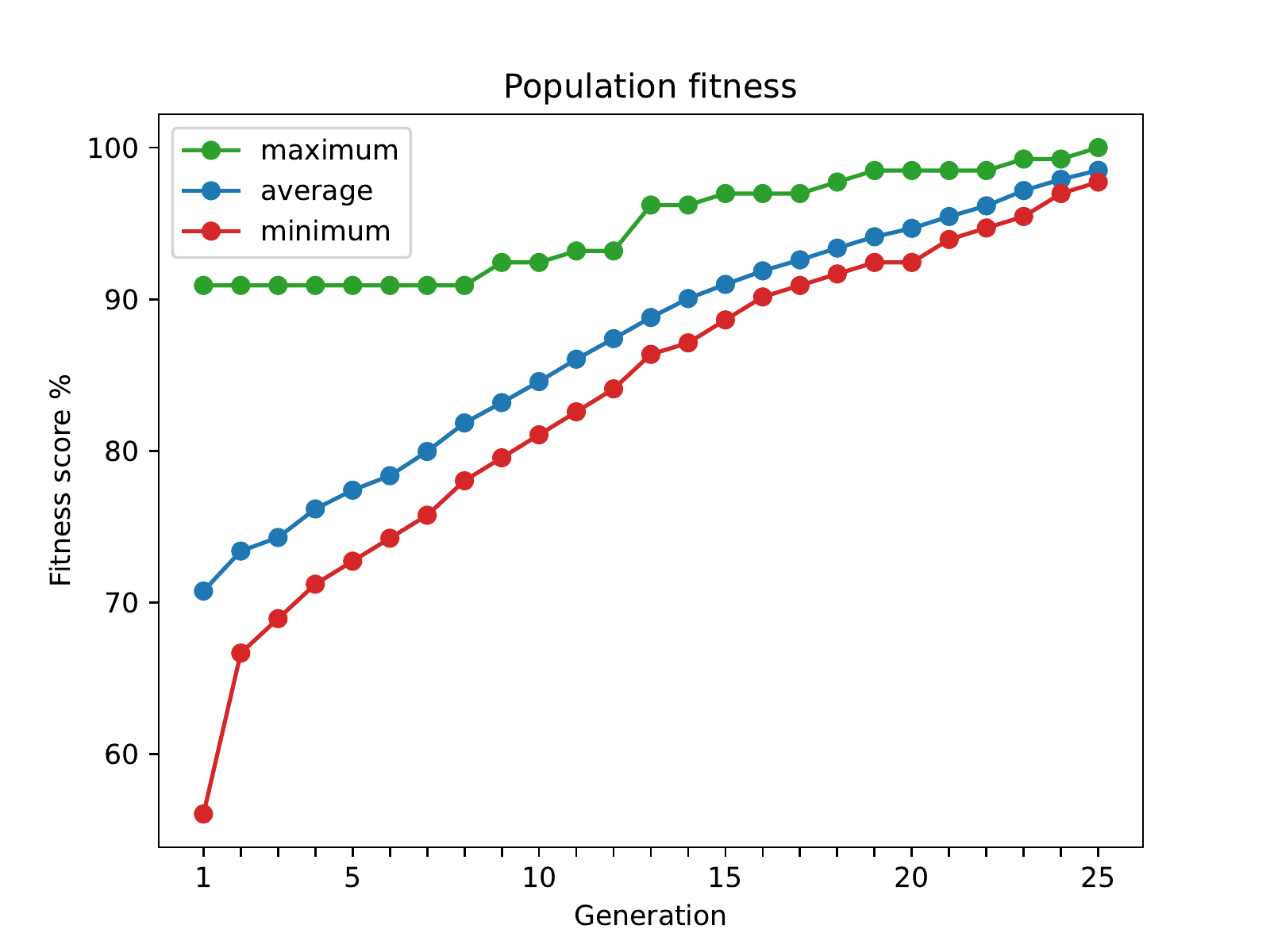}}
  
  \subfloat[Mammographic Mass]{\includegraphics[width=0.5\textwidth, height=3cm]{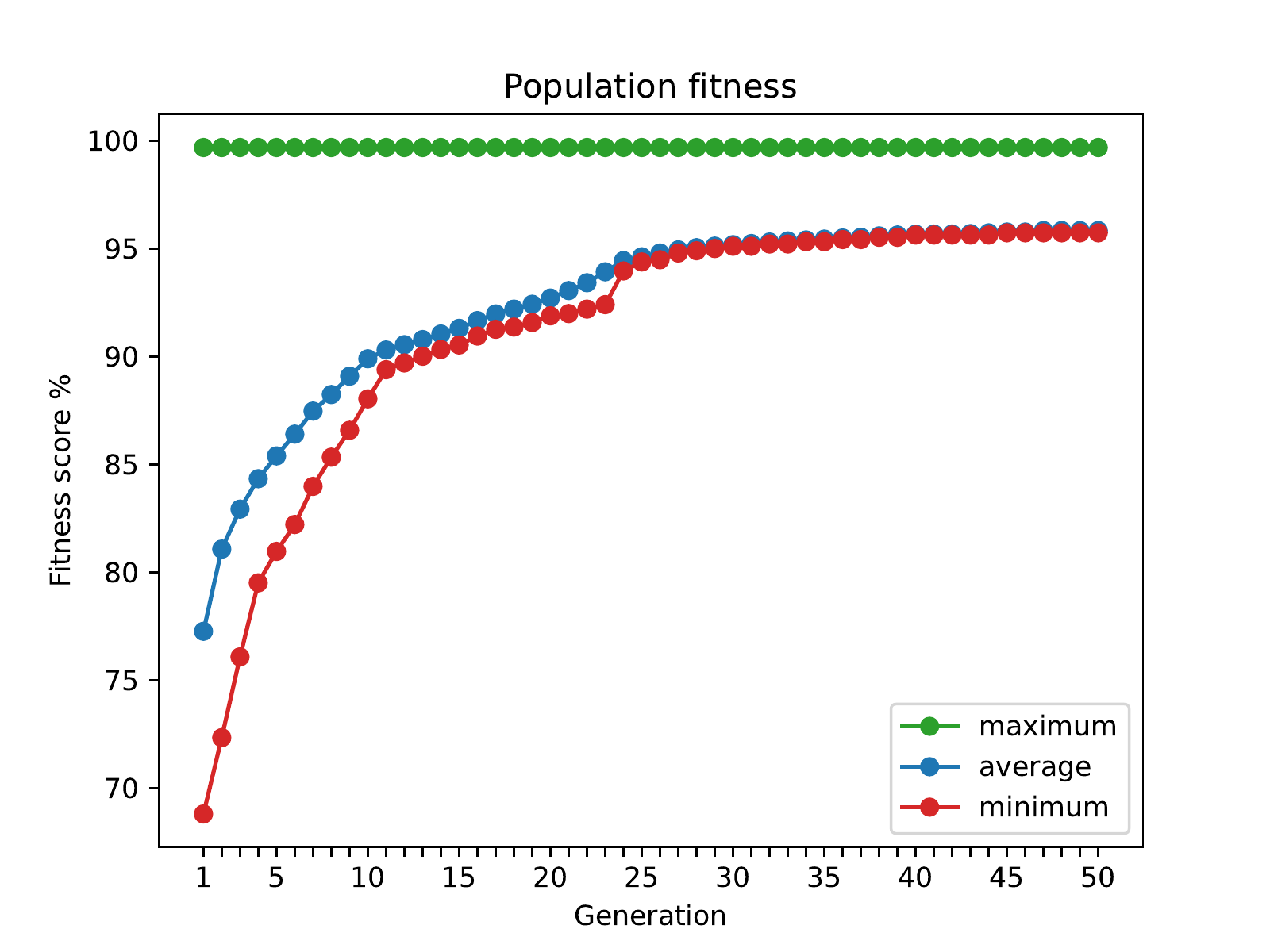}}
  \subfloat[Blood Transfusion]{\includegraphics[width=0.5\textwidth, height=3cm]{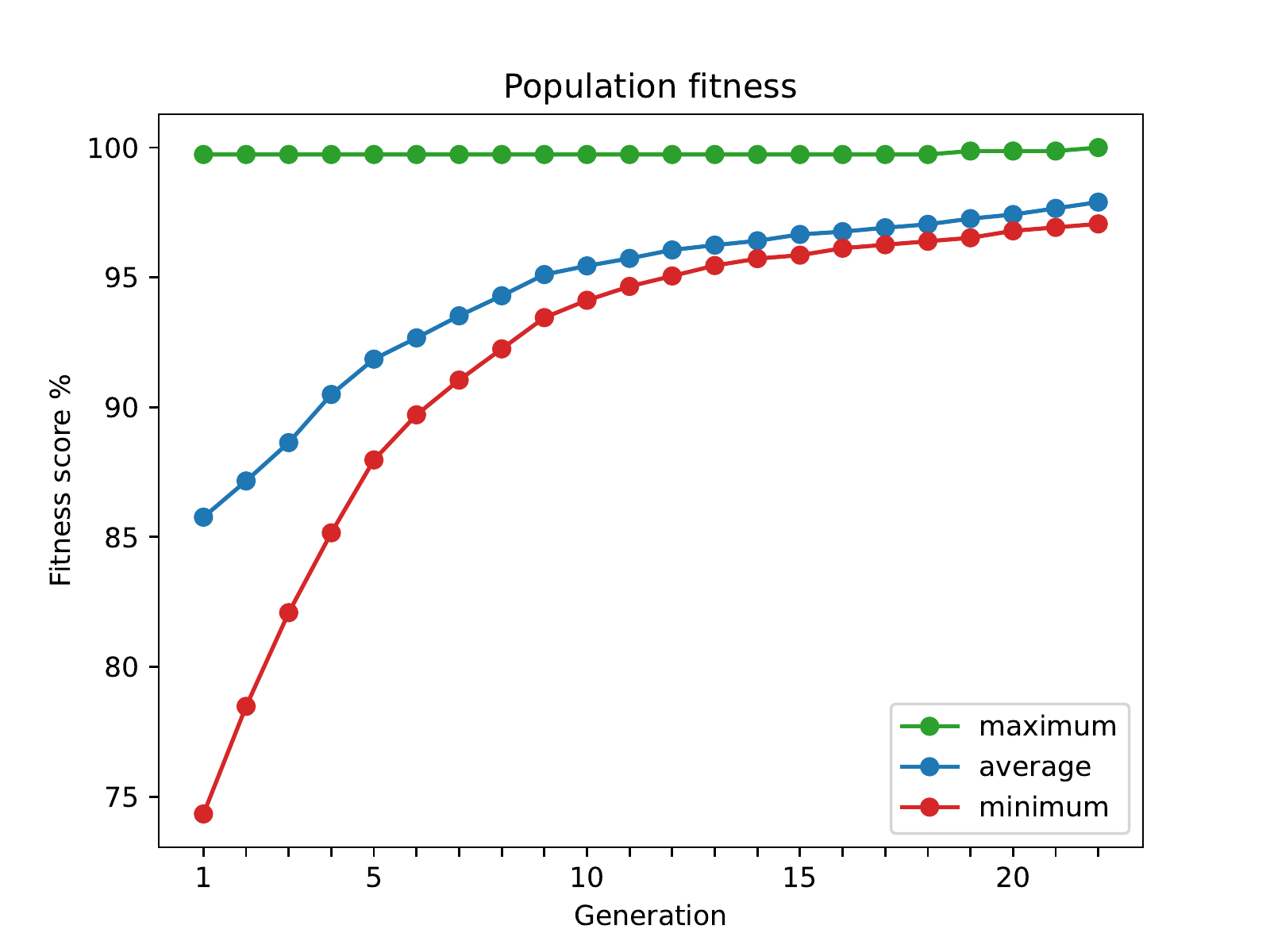}}
  
  \subfloat[Ecoli]{\includegraphics[width=0.5\textwidth, height=3cm]{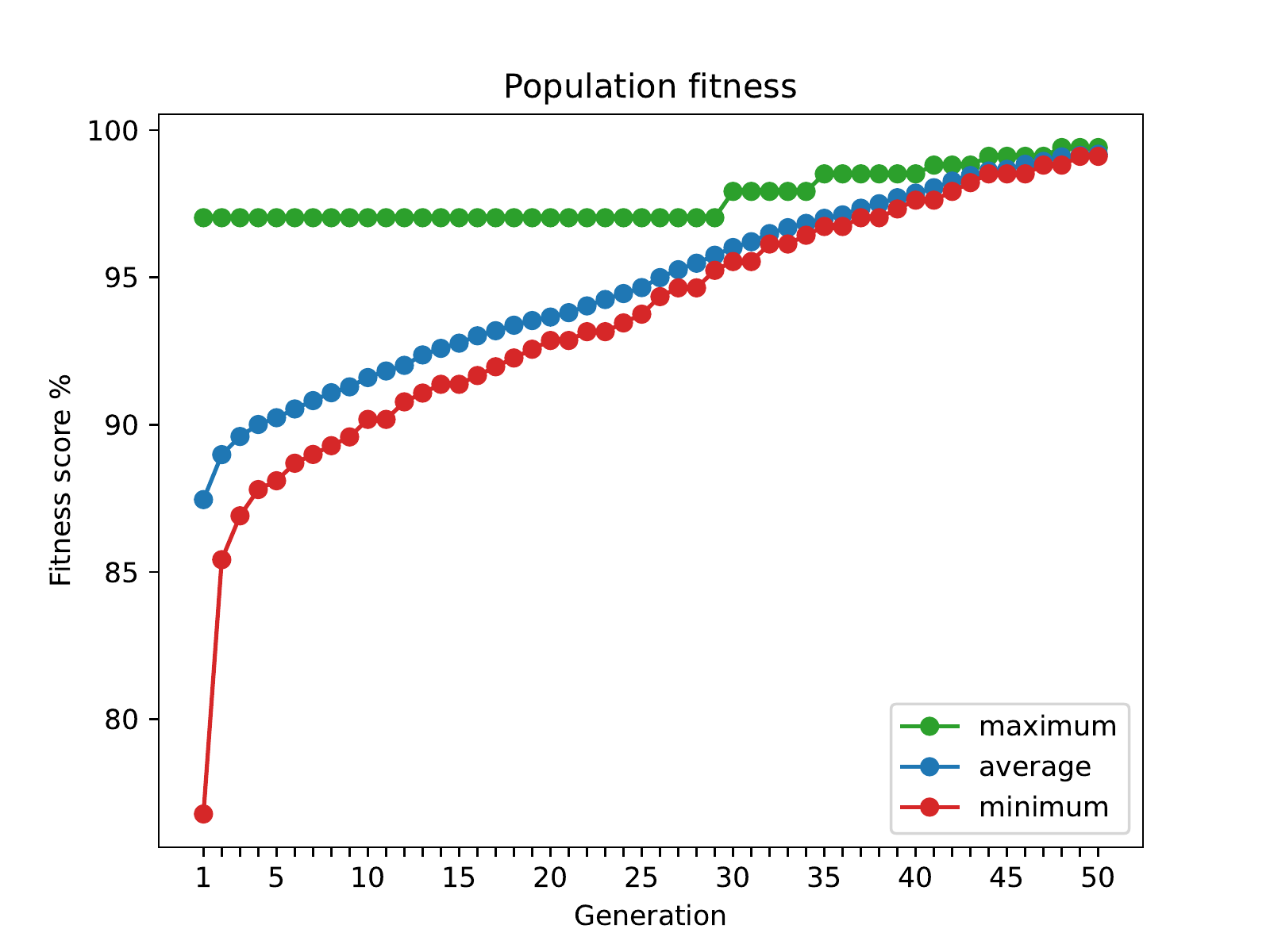}}
  \subfloat[Tic-tac-toe]{\includegraphics[width=0.5\textwidth, height=3cm]{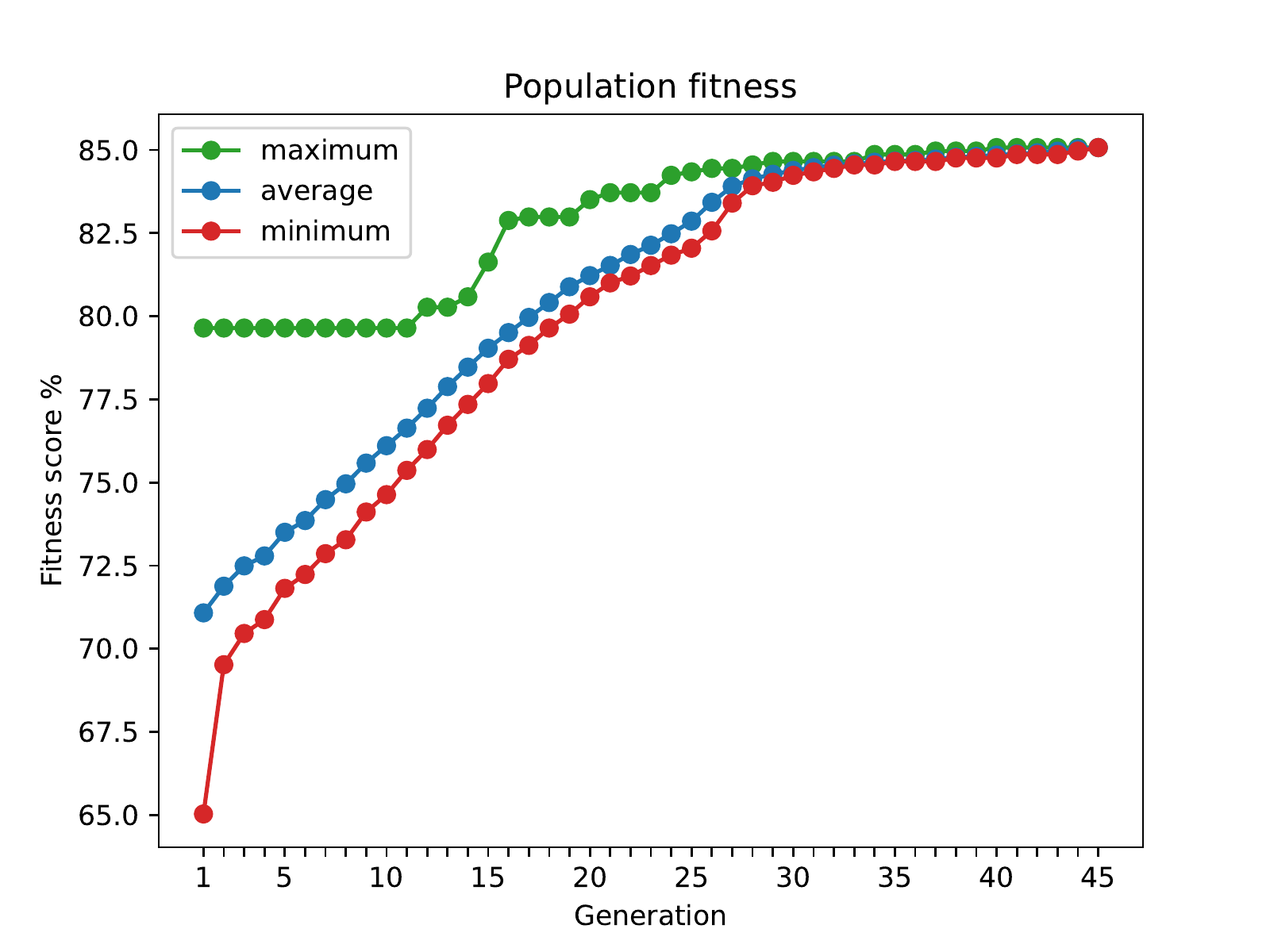}}
  \caption{Evolution of the fitness score with the quadratic genetic algorithm.}
  \label{fig:evolution}
\end{figure}

For each of the data sets, we show in Figure \ref{fig:evolution} the evolution of the fitness score of the population in each generation. For each population and generation, we keep track of the minimum, average and maximum fitness of individuals in the population. All curves show an increase over time until termination (when the maximum number of generations is reached, when the minimum and maximum fitness are equal or when the maximum fitness score has reached $100\%$).  

\section{Explanations of the Data Sets}

In Appendix \ref{sec:explanations}, we introduce each data set and show the rule set with highest coverage score obtained out of the 30 runs of the nested genetic algorithm QGA.

\section{Conclusion}
\label{section-conclusion}

Our goal was to explain classification data sets. We have formulated the problem of selecting an optimal rule set as a budgeted maximum coverage problem. This problem is chalenging, because it involves searching over exponentially many possible sets of rules. We have designed a nested genetic algorithm baptised Quadratic Genetic Algorithm to solve the problem and have used the RF+HC method to efficiently compute an initial population, which has then been expanded using an inner genetic algorithm. More importantly, a second inner genetic algorithm has allowed us to find mutations that enable the evolutionary process. We have achieved high coverage scores even with a budget for complexity and errors lower than what the best decision tree requires. We have provided explanations for 10 public classification data sets and demonstrated that the genetic algorithm offered a scalable approach to this problem.

\bibliography{sn-bibliography}


\begin{thebibliography}{28}
\ifx \bisbn   \undefined \def \bisbn  #1{ISBN #1}\fi
\ifx \binits  \undefined \def \binits#1{#1}\fi
\ifx \bauthor  \undefined \def \bauthor#1{#1}\fi
\ifx \batitle  \undefined \def \batitle#1{#1}\fi
\ifx \bjtitle  \undefined \def \bjtitle#1{#1}\fi
\ifx \bvolume  \undefined \def \bvolume#1{\textbf{#1}}\fi
\ifx \byear  \undefined \def \byear#1{#1}\fi
\ifx \bissue  \undefined \def \bissue#1{#1}\fi
\ifx \bfpage  \undefined \def \bfpage#1{#1}\fi
\ifx \blpage  \undefined \def \blpage #1{#1}\fi
\ifx \burl  \undefined \def \burl#1{\textsf{#1}}\fi
\ifx \doiurl  \undefined \def \doiurl#1{\url{https://doi.org/#1}}\fi
\ifx \betal  \undefined \def \betal{\textit{et al.}}\fi
\ifx \binstitute  \undefined \def \binstitute#1{#1}\fi
\ifx \binstitutionaled  \undefined \def \binstitutionaled#1{#1}\fi
\ifx \bctitle  \undefined \def \bctitle#1{#1}\fi
\ifx \beditor  \undefined \def \beditor#1{#1}\fi
\ifx \bpublisher  \undefined \def \bpublisher#1{#1}\fi
\ifx \bbtitle  \undefined \def \bbtitle#1{#1}\fi
\ifx \bedition  \undefined \def \bedition#1{#1}\fi
\ifx \bseriesno  \undefined \def \bseriesno#1{#1}\fi
\ifx \blocation  \undefined \def \blocation#1{#1}\fi
\ifx \bsertitle  \undefined \def \bsertitle#1{#1}\fi
\ifx \bsnm \undefined \def \bsnm#1{#1}\fi
\ifx \bsuffix \undefined \def \bsuffix#1{#1}\fi
\ifx \bparticle \undefined \def \bparticle#1{#1}\fi
\ifx \barticle \undefined \def \barticle#1{#1}\fi
\bibcommenthead
\ifx \bconfdate \undefined \def \bconfdate #1{#1}\fi
\ifx \botherref \undefined \def \botherref #1{#1}\fi
\ifx \url \undefined \def \url#1{\textsf{#1}}\fi
\ifx \bchapter \undefined \def \bchapter#1{#1}\fi
\ifx \bbook \undefined \def \bbook#1{#1}\fi
\ifx \bcomment \undefined \def \bcomment#1{#1}\fi
\ifx \oauthor \undefined \def \oauthor#1{#1}\fi
\ifx \citeauthoryear \undefined \def \citeauthoryear#1{#1}\fi
\ifx \endbibitem  \undefined \def \endbibitem {}\fi
\ifx \bconflocation  \undefined \def \bconflocation#1{#1}\fi
\ifx \arxivurl  \undefined \def \arxivurl#1{\textsf{#1}}\fi
\csname PreBibitemsHook\endcsname

\bibitem{samuel1959some}
\begin{botherref}
\oauthor{\bsnm{Samuel}, \binits{A.}}:
Some studies in machine learning using the game of checkers. Reprinted in EA
  Feigenbaum \& J. Feldman (Eds.)(1963). Computers and thought.
McGraw-Hill
(1959)
\end{botherref}
\endbibitem

\bibitem{lecun2015deep}
\begin{barticle}
\bauthor{\bsnm{LeCun}, \binits{Y.}},
\bauthor{\bsnm{Bengio}, \binits{Y.}},
\bauthor{\bsnm{Hinton}, \binits{G.}}:
\batitle{Deep learning}.
\bjtitle{nature}
\bvolume{521}(\bissue{7553}),
\bfpage{436}--\blpage{444}
(\byear{2015})
\end{barticle}
\endbibitem

\bibitem{guidotti2018survey}
\begin{barticle}
\bauthor{\bsnm{Guidotti}, \binits{R.}},
\bauthor{\bsnm{Monreale}, \binits{A.}},
\bauthor{\bsnm{Ruggieri}, \binits{S.}},
\bauthor{\bsnm{Turini}, \binits{F.}},
\bauthor{\bsnm{Giannotti}, \binits{F.}},
\bauthor{\bsnm{Pedreschi}, \binits{D.}}:
\batitle{A survey of methods for explaining black box models}.
\bjtitle{ACM computing surveys (CSUR)}
\bvolume{51}(\bissue{5}),
\bfpage{1}--\blpage{42}
(\byear{2018})
\end{barticle}
\endbibitem

\bibitem{molnar2020interpretable}
\begin{botherref}
\oauthor{\bsnm{Molnar}, \binits{C.}}:
Interpretable machine learning.
Lulu. com
(2020)
\end{botherref}
\endbibitem

\bibitem{samek2019explainable}
\begin{botherref}
\oauthor{\bsnm{Samek}, \binits{W.}},
\oauthor{\bsnm{Montavon}, \binits{G.}},
\oauthor{\bsnm{Vedaldi}, \binits{A.}},
\oauthor{\bsnm{Hansen}, \binits{L.K.}},
\oauthor{\bsnm{M{\"u}ller}, \binits{K.-R.}}:
Explainable AI: interpreting, explaining and visualizing deep learning.
Springer
(2019)
\end{botherref}
\endbibitem

\bibitem{xu2019explainable}
\begin{bchapter}
\bauthor{\bsnm{Xu}, \binits{F.}},
\bauthor{\bsnm{Uszkoreit}, \binits{H.}},
\bauthor{\bsnm{Du}, \binits{Y.}},
\bauthor{\bsnm{Fan}, \binits{W.}},
\bauthor{\bsnm{Zhao}, \binits{D.}},
\bauthor{\bsnm{Zhu}, \binits{J.}}:
\bctitle{Explainable ai: A brief survey on history, research areas, approaches
  and challenges}.
In: \bbtitle{CCF International Conference on Natural Language Processing and
  Chinese Computing},
pp. \bfpage{563}--\blpage{574}
(\byear{2019}).
\bcomment{Springer}
\end{bchapter}
\endbibitem

\bibitem{burkart2021survey}
\begin{barticle}
\bauthor{\bsnm{Burkart}, \binits{N.}},
\bauthor{\bsnm{Huber}, \binits{M.F.}}:
\batitle{A survey on the explainability of supervised machine learning}.
\bjtitle{Journal of Artificial Intelligence Research}
\bvolume{70},
\bfpage{245}--\blpage{317}
(\byear{2021})
\end{barticle}
\endbibitem

\bibitem{gunning2019xai}
\begin{barticle}
\bauthor{\bsnm{Gunning}, \binits{D.}},
\bauthor{\bsnm{Stefik}, \binits{M.}},
\bauthor{\bsnm{Choi}, \binits{J.}},
\bauthor{\bsnm{Miller}, \binits{T.}},
\bauthor{\bsnm{Stumpf}, \binits{S.}},
\bauthor{\bsnm{Yang}, \binits{G.-Z.}}:
\batitle{Xai—explainable artificial intelligence}.
\bjtitle{Science robotics}
\bvolume{4}(\bissue{37}),
\bfpage{7120}
(\byear{2019})
\end{barticle}
\endbibitem

\bibitem{vale2022explainable}
\begin{botherref}
\oauthor{\bsnm{Vale}, \binits{D.}},
\oauthor{\bsnm{El-Sharif}, \binits{A.}},
\oauthor{\bsnm{Ali}, \binits{M.}}:
Explainable artificial intelligence (xai) post-hoc explainability methods:
  Risks and limitations in non-discrimination law.
AI and Ethics,
1--12
(2022)
\end{botherref}
\endbibitem

\bibitem{rudin2019stop}
\begin{barticle}
\bauthor{\bsnm{Rudin}, \binits{C.}}:
\batitle{Stop explaining black box machine learning models for high stakes
  decisions and use interpretable models instead}.
\bjtitle{Nature Machine Intelligence}
\bvolume{1}(\bissue{5}),
\bfpage{206}--\blpage{215}
(\byear{2019})
\end{barticle}
\endbibitem

\bibitem{vasic2019moet}
\begin{botherref}
\oauthor{\bsnm{Vasic}, \binits{M.}},
\oauthor{\bsnm{Petrovic}, \binits{A.}},
\oauthor{\bsnm{Wang}, \binits{K.}},
\oauthor{\bsnm{Nikolic}, \binits{M.}},
\oauthor{\bsnm{Singh}, \binits{R.}},
\oauthor{\bsnm{Khurshid}, \binits{S.}}:
Mo{\"e}t: Interpretable and verifiable reinforcement learning via mixture of
  expert trees
(2019)
\end{botherref}
\endbibitem

\bibitem{wang2017bayesian}
\begin{barticle}
\bauthor{\bsnm{Wang}, \binits{T.}},
\bauthor{\bsnm{Rudin}, \binits{C.}},
\bauthor{\bsnm{Doshi-Velez}, \binits{F.}},
\bauthor{\bsnm{Liu}, \binits{Y.}},
\bauthor{\bsnm{Klampfl}, \binits{E.}},
\bauthor{\bsnm{MacNeille}, \binits{P.}}:
\batitle{A bayesian framework for learning rule sets for interpretable
  classification}.
\bjtitle{The Journal of Machine Learning Research}
\bvolume{18}(\bissue{1}),
\bfpage{2357}--\blpage{2393}
(\byear{2017})
\end{barticle}
\endbibitem

\bibitem{khuller1999budgeted}
\begin{barticle}
\bauthor{\bsnm{Khuller}, \binits{S.}},
\bauthor{\bsnm{Moss}, \binits{A.}},
\bauthor{\bsnm{Naor}, \binits{J.S.}}:
\batitle{The budgeted maximum coverage problem}.
\bjtitle{Information processing letters}
\bvolume{70}(\bissue{1}),
\bfpage{39}--\blpage{45}
(\byear{1999})
\end{barticle}
\endbibitem

\bibitem{panchal2015solving}
\begin{barticle}
\bauthor{\bsnm{Panchal}, \binits{G.}},
\bauthor{\bsnm{Panchal}, \binits{D.}}:
\batitle{Solving {NP} hard problems using genetic algorithm}.
\bjtitle{Transportation}
\bvolume{106},
\bfpage{6}--\blpage{2}
(\byear{2015})
\end{barticle}
\endbibitem

\bibitem{AdnanIslam2017forex}
\begin{botherref}
\oauthor{\bsnm{Adnan}, \binits{N.}},
\oauthor{\bsnm{Islam}, \binits{Z.}}:
Forex++: A new framework for knowledge discovery from decision forests.
Australasian Journal of Information Systems
\textbf{21}
(2017).
\doiurl{10.3127/ajis.v21i0.1539}
\end{botherref}
\endbibitem

\bibitem{gras2015rfhc}
\begin{bchapter}
\bauthor{\bsnm{Mashayekhi}, \binits{M.}},
\bauthor{\bsnm{Gras}, \binits{R.}}:
\bctitle{Rule extraction from random forest: the rf+hc methods}.
In: \beditor{\bsnm{Barbosa}, \binits{D.}},
\beditor{\bsnm{Milios}, \binits{E.}} (eds.)
\bbtitle{Advances in Artificial Intelligence},
pp. \bfpage{223}--\blpage{237}.
\bpublisher{Springer},
\blocation{Cham}
(\byear{2015})
\end{bchapter}
\endbibitem

\bibitem{wang2020irfre}
\begin{barticle}
\bauthor{\bsnm{Wang}, \binits{S.}},
\bauthor{\bsnm{Wang}, \binits{Y.}},
\bauthor{\bsnm{Wang}, \binits{D.}},
\bauthor{\bsnm{Yin}, \binits{Y.}},
\bauthor{\bsnm{Wang}, \binits{Y.}},
\bauthor{\bsnm{Jin}, \binits{Y.}}:
\batitle{An improved random forest-based rule extraction method for breast
  cancer diagnosis}.
\bjtitle{Applied Soft Computing}
\bvolume{86},
\bfpage{105941}
(\byear{2020}).
\doiurl{10.1016/j.asoc.2019.105941}
\end{barticle}
\endbibitem

\bibitem{fonseca1993multiobjective}
\begin{bchapter}
\bauthor{\bsnm{Fonseca}, \binits{C.M.}},
\bauthor{\bsnm{Fleming}, \binits{P.J.}}:
\bctitle{Multiobjective genetic algorithms}.
In: \bbtitle{IEE Colloquium on Genetic Algorithms for Control Systems
  Engineering},
pp. \bfpage{6}--\blpage{1}
(\byear{1993}).
\bcomment{Iet}
\end{bchapter}
\endbibitem

\bibitem{izza2020explaining}
\begin{botherref}
\oauthor{\bsnm{Izza}, \binits{Y.}},
\oauthor{\bsnm{Ignatiev}, \binits{A.}},
\oauthor{\bsnm{Marques-Silva}, \binits{J.}}:
On explaining decision trees.
arXiv preprint arXiv:2010.11034
(2020)
\end{botherref}
\endbibitem

\bibitem{izza2022tackling}
\begin{botherref}
\oauthor{\bsnm{Izza}, \binits{Y.}},
\oauthor{\bsnm{Ignatiev}, \binits{A.}},
\oauthor{\bsnm{Marques-Silva}, \binits{J.}}:
On tackling explanation redundancy in decision trees.
arXiv preprint arXiv:2205.09971
(2022)
\end{botherref}
\endbibitem

\bibitem{verhaeghe2020learning}
\begin{barticle}
\bauthor{\bsnm{Verhaeghe}, \binits{H.}},
\bauthor{\bsnm{Nijssen}, \binits{S.}},
\bauthor{\bsnm{Pesant}, \binits{G.}},
\bauthor{\bsnm{Quimper}, \binits{C.-G.}},
\bauthor{\bsnm{Schaus}, \binits{P.}}:
\batitle{Learning optimal decision trees using constraint programming}.
\bjtitle{Constraints}
\bvolume{25}(\bissue{3}),
\bfpage{226}--\blpage{250}
(\byear{2020})
\end{barticle}
\endbibitem

\bibitem{alos2021learning}
\begin{botherref}
\oauthor{\bsnm{Alos}, \binits{J.}},
\oauthor{\bsnm{Ansotegui}, \binits{C.}},
\oauthor{\bsnm{Torres}, \binits{E.}}:
Learning optimal decision trees using maxsat.
arXiv preprint arXiv:2110.13854
(2021)
\end{botherref}
\endbibitem

\bibitem{demirovic2022murtree}
\begin{barticle}
\bauthor{\bsnm{Demirovi{\'c}}, \binits{E.}},
\bauthor{\bsnm{Lukina}, \binits{A.}},
\bauthor{\bsnm{Hebrard}, \binits{E.}},
\bauthor{\bsnm{Chan}, \binits{J.}},
\bauthor{\bsnm{Bailey}, \binits{J.}},
\bauthor{\bsnm{Leckie}, \binits{C.}},
\bauthor{\bsnm{Ramamohanarao}, \binits{K.}},
\bauthor{\bsnm{Stuckey}, \binits{P.J.}}:
\batitle{Murtree: Optimal decision trees via dynamic programming and search}.
\bjtitle{Journal of Machine Learning Research}
\bvolume{23}(\bissue{26}),
\bfpage{1}--\blpage{47}
(\byear{2022})
\end{barticle}
\endbibitem

\bibitem{lakkaraju2016interpretable}
\begin{bchapter}
\bauthor{\bsnm{Lakkaraju}, \binits{H.}},
\bauthor{\bsnm{Bach}, \binits{S.H.}},
\bauthor{\bsnm{Leskovec}, \binits{J.}}:
\bctitle{Interpretable decision sets: A joint framework for description and
  prediction}.
In: \bbtitle{Proceedings of the 22nd ACM SIGKDD International Conference on
  Knowledge Discovery and Data Mining},
pp. \bfpage{1675}--\blpage{1684}
(\byear{2016})
\end{bchapter}
\endbibitem

\bibitem{borgelt2005implementation}
\begin{bchapter}
\bauthor{\bsnm{Borgelt}, \binits{C.}}:
\bctitle{An implementation of the {FP}-growth algorithm}.
In: \bbtitle{Proceedings of the 1st International Workshop on Open Source Data
  Mining: Frequent Pattern Mining Implementations},
pp. \bfpage{1}--\blpage{5}
(\byear{2005})
\end{bchapter}
\endbibitem

\bibitem{goldberg1989genetic}
\begin{botherref}
\oauthor{\bsnm{Goldberg}, \binits{D.E.}}:
Genetic algorithms in search, optimization, and machine learning. addison.
Reading
(1989)
\end{botherref}
\endbibitem

\bibitem{Dua:2019}
\begin{botherref}
\oauthor{\bsnm{Dua}, \binits{D.}},
\oauthor{\bsnm{Graff}, \binits{C.}}:
{UCI} Machine Learning Repository
(2017).
\url{http://archive.ics.uci.edu/ml}
\end{botherref}
\endbibitem

\bibitem{hayes1977concept}
\begin{barticle}
\bauthor{\bsnm{Hayes-Roth}, \binits{B.}},
\bauthor{\bsnm{Hayes-Roth}, \binits{F.}}:
\batitle{Concept learning and the recognition and classification of exemplars}.
\bjtitle{Journal of Verbal Learning and Verbal Behavior}
\bvolume{16}(\bissue{3}),
\bfpage{321}--\blpage{338}
(\byear{1977})
\end{barticle}
\endbibitem

\end{thebibliography}

\newpage
\begin{appendices}

\section{Explanation of the Data Sets}
\label{sec:explanations}

\subsubsection{Iris Data Set}

This is perhaps the best known data set to be found in the pattern recognition literature. The data set contains 3 classes of 50 instances each, where each class refers to a type of iris plant: setosa, versicolor or virginica. There are 4 continuous attributes: 
\begin{itemize}
    \item sepal length in cm
    \item sepal width in cm
    \item petal length in cm
    \item petal width in cm
\end{itemize}
The data set can be found here\footnote{\url{https://archive.ics.uci.edu/ml/datasets/iris}}. The explanation is printed in Figure \ref{explanation-iris}. It is composed of 3 rules and has a complexity of 4, $94\%$ coverage and 2 classification errors.

\begin{figure}[H]
\begin{tiny}
\begin{verbatim}
IF petal length <= 2.45 THEN CLASS=Iris-setosa
IF petal width > 0.8 AND petal length <= 4.75 THEN CLASS=Iris-versicolor
IF petal width > 1.75 THEN CLASS=Iris-virginica

\end{verbatim}
\caption{Explanation of the Iris data set.}
\label{explanation-iris}
\end{tiny}
\end{figure}

\subsubsection{Wine Data Set}

These data are the results of a chemical analysis of 178 wines grown in the same region in Italy but derived from 3 different cultivars. The analysis determined the quantities of 12 constituents found in each of the three types of wines. The 12 attributes are:
\begin{itemize}
    \item Malic acid
    \item Ash
    \item Alcalinity of ash
    \item Magnesium
    \item Total phenols
    \item Flavanoids
    \item Nonflavanoid phenols
    \item Proanthocyanins
    \item Color intensity
    \item Hue
    \item OD280/OD315 of diluted wines
    \item Proline 
\end{itemize}
The class is an integer that represents the cultivar. The data set can be found here\footnote{\url{https://archive.ics.uci.edu/ml/datasets/wine}}. The explanation is printed in Figure \ref{explanation-wine}. It is composed of 3 rules, has a complexity of 8, a coverage of $96.1\%$ and $8$ classification errors.

\begin{figure}[H]
\begin{tiny}
\begin{verbatim}
IF hue > 0.785 AND color intensity > 3.46 AND flavanoids > 2.11 THEN CLASS=1
IF color intensity > 3.82 AND flavanoids <= 1.58 AND alcalinity > 17.15 THEN CLASS=3
IF color intensity <= 3.82 AND OD280/OD135 <= 3.73 THEN CLASS=2

\end{verbatim}
\caption{Explanation of the Wine data set.}
\label{explanation-wine}
\end{tiny}
\end{figure}

\subsubsection{Zoo Data Set}

This is a simple data set containing 101 animals split in 7 classes. Here is a breakdown of the animals in each class:
\begin{enumerate}
    \item Aardvark, antelope, bear, boar, buffalo, calf, cavy, cheetah, deer, dolphin, elephant, fruitbat, giraffe, girl, goat, gorilla, hamster, hare, leopard, lion, lynx, mink, mole, mongoose, opossum, oryx, platypus, polecat, pony, porpoise, puma, pussycat, raccoon, reindeer, seal, sealion, squirrel, vampire, vole, wallaby, wolf
    \item Chicken, crow, dove, duck, flamingo, gull, hawk, kiwi, lark, ostrich, parakeet, penguin, pheasant, rhea, skimmer, skua, sparrow, swan, vulture, wren
    \item Pitviper, seasnake, slowworm, tortoise, tuatara
    \item Bass, carp, catfish, chub, dogfish, haddock, herring, pike, piranha, seahorse, sole, stingray, tuna
    \item Frog, newt, toad
    \item Flea, gnat, honeybee, housefly, ladybird, moth, termite, wasp
    \item Clam, crab, crayfish, lobster, octopus, scorpion, seawasp, slug, starfish, worm
\end{enumerate}
The 17 attributes are:
\begin{itemize}
    \item hair: Boolean
    \item feathers: Boolean
    \item eggs: Boolean
    \item milk: Boolean
    \item airborne: Boolean
    \item aquatic: Boolean
    \item predator: Boolean
    \item toothed: Boolean
    \item backbone: Boolean
    \item breathes: Boolean
    \item venomous: Boolean
    \item fins: Boolean
    \item legs: 0, 2, 4, 6, 8
    \item tail: Boolean
    \item domestic: Boolean
    \item catsize: Boolean
\end{itemize}

The data set is described and can be downloaded here\footnote{\url{https://archive.ics.uci.edu/ml/datasets/zoo}}. The explanation is printed in Figure \ref{explanation-zoo}. It is composed of 7 rules, has a complexity of 18, a coverage of 100\% and 8 classification errors.

\begin{figure}[H]
\begin{tiny}
\begin{verbatim}
IF milk <= 0.5 AND feathers > 0.5 THEN CLASS=2
IF milk > 0.5 THEN CLASS=1
IF hair > 0.5 AND milk <= 0.5 THEN CLASS=6
IF milk <= 0.032 AND fins <= 0.764 AND breathes > 0.205 AND toothed > 0.563 THEN CLASS=5
IF breathes <= 0.371 AND backbone <= 0.101 THEN CLASS=7
IF toothed > 0.5 AND hair <= 0.5 AND breathes <= 0.5 THEN CLASS=4
IF milk <= 0.5 AND feathers <= 0.5 AND toothed <= 0.5 AND breathes > 0.5 THEN CLASS=6

\end{verbatim}
\caption{Explanation of the Zoo data set.}
\label{explanation-zoo}
\end{tiny}
\end{figure}

\subsubsection{Breast Cancer Wisconsin (Original) Data Set}

This data set is composed of 699 clinical exams of cells (cytology) used for breast cancer diagnosis. The 9 attributes are the following cytological characteristics:
\begin{itemize}
    \item Clump Thickness: 1 - 10
    \item Uniformity of Cell Size: 1 - 10
    \item Uniformity of Cell Shape: 1 - 10
    \item Marginal Adhesion: 1 - 10
    \item Single Epithelial Cell Size: 1 - 10
    \item Bare Nuclei: 1 - 10
    \item Bland Chromatin: 1 - 10
    \item Normal Nucleoli: 1 - 10
    \item Mitoses: 1 - 10
\end{itemize}

The 2 classes are 2=\enquote{benign} and 4=\enquote{malignant}.

The data set is described and can be downloaded here\footnote{\url{https://archive.ics.uci.edu/ml/datasets/breast+cancer+wisconsin+(original)}}. The explanation is printed in Figure \ref{explanation-breast-cancer}.
The explanation is composed of 8 rules, has a complexity of 16, a coverage of $98.9\%$ and 28 classification errors.

\begin{figure}[H]
\begin{tiny}
\begin{verbatim}
IF bland chromatin <= 3.5 AND clump thickness <= 6.5 THEN CLASS=2
IF bland chromatin > 7.5 THEN CLASS=4
IF clump thickness > 6.409 THEN CLASS=4
IF clump thickness <= 4.612 AND adhesion <= 1.843 THEN CLASS=2
IF bland chromatin > 3.5 AND uniformity cell size > 4.5 THEN CLASS=4
IF bland chromatin > 3.5 AND clump thickness > 4.5 AND uniformity cell size <= 4.5 THEN CLASS=4
IF bland chromatin > 3.5 AND bare nuclei_10 > 0.5 THEN CLASS=4
IF uniformity cell shape <= 2.5 AND bland chromatin > 3.5 AND bare nuclei_1 > 0.5 THEN CLASS=2

\end{verbatim}
\caption{Explanation of the Breast cancer Wisconsin data set.}
\label{explanation-breast-cancer}
\end{tiny}
\end{figure}

\subsubsection{Banknote Authentication Data Set}

Data were extracted from images that were taken from 1372 genuine and forged banknote-like specimens. For digitization, an industrial camera usually used for print inspection was used. The final images have 400x400 pixels. Due to the object lens and distance to the investigated object gray-scale pictures with a resolution of about 660 dpi were gained. Wavelet Transform tool were used to extract features from images. The 4 attributes are:
\begin{itemize}
    \item variance of Wavelet Transformed image (continuous)
    \item skewness of Wavelet Transformed image (continuous)
    \item curtosis of Wavelet Transformed image (continuous)
    \item entropy of image (continuous)
\end{itemize}
The class is a Boolean describing if the banknotes are authentic or not. The data set is described and can be downloaded here\footnote{\url{https://archive.ics.uci.edu/ml/datasets/banknote+authentication}}. The explanation is printed in Figure \ref{explanation-banknote}. The explanation is composed of $10$ rules, has a complexity of $23$, a coverage of $99.6\%$ and $33$ errors.

\begin{figure}[H]
\begin{tiny}
\centering
\begin{verbatim}
IF skewness > 5.21 AND variance > -3.368 THEN CLASS=0
IF skewness > 5.161 AND variance > -3.368 THEN CLASS=0
IF skewness <= 5.161 AND variance > 0.32 AND curtosis <= -1.44 THEN CLASS=1
IF skewness <= 5.161 AND variance > 0.32 AND curtosis > -1.44 THEN CLASS=0
IF curtosis > 2.448 AND variance <= 0.323 AND skewness <= 0.667 THEN CLASS=1
IF curtosis > 2.448 AND variance <= 0.323 AND skewness > 0.667 THEN CLASS=0
IF curtosis > 2.448 AND variance > 0.323 THEN CLASS=0
IF variance <= -4.417 THEN CLASS=1
IF variance <= 0.32 AND skewness <= 5.865 AND curtosis <= 6.219 THEN CLASS=1
IF variance > 6.079 THEN CLASS=0

\end{verbatim}
\caption{Explanation of the Banknote authentication data set.}
\label{explanation-banknote}
\end{tiny}
\end{figure}

\subsubsection{Hayes-Roth Data Set}

This data set contains 160 instances split into 3 classes. It has 3 attributes.
\begin{itemize}
    \item age: categorical values ranging between 1 and 4 (30, 40, 50, 60 years old)
    \item educational level: categorical values ranging between 1 and 4 (junior high, high school, trade school, college)
    \item marital status: categorical values ranging between 1 and 4 (single, married, divorced, widowed)
\end{itemize}
The class is an integer between 1 and 3 representing the membership of individuals that belong to Club 1, Club 2 or neither. The data set \cite{hayes1977concept} is described and can be downloaded here\footnote{\url{https://archive.ics.uci.edu/ml/datasets/Hayes-Roth}}. The explanation is printed in Figure \ref{explanation-hayes-roth}. The explanation is composed of $17$ rules, has a complexity of $67$, a coverage of $100\%$ and $14$ classification errors.

\begin{figure}[H]
\begin{tiny}
\centering
\begin{verbatim}
IF educational level > 3.5 THEN CLASS=3
IF age > 3.5 THEN CLASS=3
IF age <= 3.5 AND educational level > 3.5 THEN CLASS=3
IF age <= 3.5 AND marital status > 3.5 THEN CLASS=3
IF age <= 3.5 AND educational level <= 3.5 AND marital status <= 3.5 AND marital status > 1.5 AND age > 1.5 THEN CLASS=2
IF age <= 3.5 AND educational level <= 3.5 AND marital status <= 3.5 AND educational level <= 1.5 AND marital status <= 1.5 
THEN CLASS=1
IF age <= 1.5 AND educational level <= 3.5 AND marital status <= 3.5 AND educational level <= 1.5 AND marital status > 1.5 
THEN CLASS=1
IF marital status > 3.5 THEN CLASS=3
IF marital status <= 3.5 AND name > 117.5 AND educational level > 2.5 THEN CLASS=1
IF name <= 88.5 AND marital status <= 3.5 AND marital status <= 1.5 AND name > 82.0 AND educational level <= 3.0 THEN CLASS=2
IF marital status <= 1.618 AND educational level > 2.35 AND hobby > 1.1 AND educational level <= 3.077 AND name <= 106.44 
THEN CLASS=1
IF marital status > 2.123 AND age <= 1.397 AND hobby <= 2.969 AND marital status <= 3.321 AND name > 69.613 THEN CLASS=1
IF educational level <= 2.541 AND age <= 3.204 AND marital status <= 1.071 AND educational level > 1.123 AND age <= 1.31 
THEN CLASS=1
IF educational level <= 2.541 AND age <= 3.204 AND marital status <= 1.071 AND educational level > 1.123 AND age > 1.31 
THEN CLASS=2
IF age <= 2.571 AND marital status <= 2.178 AND marital status <= 1.484 AND age > 1.34 AND name <= 99.674 AND 
educational level <= 3.926 AND educational level <= 1.61 THEN CLASS=1
IF educational level <= 3.5 AND age <= 3.5 AND name > 117.5 AND marital status > 3.5 THEN CLASS=3
IF age <= 1.5 AND marital status <= 3.5 AND marital status > 1.5 AND educational level <= 3.5 AND educational level > 1.5 AND 
name <= 117.5 THEN CLASS=2

\end{verbatim}
\caption{Explanation of the Hayes-Roth data set.}
\label{explanation-hayes-roth}
\end{tiny}
\end{figure}

\subsubsection{Mammographic Mass Data Set}

Mammography is the most effective method for breast cancer screening
available today. However, the low positive predictive value of breast
biopsy resulting from mammogram interpretation leads to approximately
70\% of unnecessary biopsies with benign outcomes. To reduce the high
number of unnecessary breast biopsies, several computer-aided diagnosis
(CAD) systems have been proposed in the last years. These systems
help physicians in their decision to perform a breast biopsy on a suspicious
lesion seen in a mammogram or to perform a short term follow-up
examination instead.
This data set can be used to predict the severity (benign or malignant)
of a mammographic mass lesion from BI-RADS attributes and the patient's age.
It contains a BI-RADS assessment, the patient's age and three BI-RADS attributes
together with the ground truth (the severity field) for 516 benign and
445 malignant masses that have been identified on full field digital mammograms
collected at the Institute of Radiology of the
University Erlangen-Nuremberg between 2003 and 2006.
Each instance has an associated BI-RADS assessment ranging from 1 (definitely benign)
to 5 (highly suggestive of malignancy) assigned in a double-review process by
physicians. Assuming that all cases with BI-RADS assessments greater or equal
a given value (varying from 1 to 5), are malignant and the other cases benign,
sensitivities and associated specificities can be calculated. These can be an
indication of how well a CAD system performs compared to the radiologists.

The 5 attributes used are:
\begin{itemize}
    \item BI-RADS assessment: 1 to 5 (ordinal, non-predictive!)
    \item Age: patient's age in years (integer)
    \item Shape: mass shape: round=1 oval=2 lobular=3 irregular=4 (categorical)
    \item Margin: mass margin: circumscribed=1 microlobulated=2 obscured=3 ill-defined=4 spiculated=5 (categorical)
    \item Density: mass density high=1 iso=2 low=3 fat-containing=4 (ordinal)
\end{itemize}
The class indicates the severity: benign=0 or malignant=1. The data set is described and can be downloaded here\footnote{\url{http://archive.ics.uci.edu/ml/datasets/mammographic+mass}}. The explanation is printed in Figure \ref{explanation-mammographic-mass}. It is composed of 16 decision rules, has a complexity of 71, a coverage of $99.69\%$ and 150 classification errors.

\begin{figure}[H]
\begin{tiny}
\centering
\begin{verbatim}
IF shape_1 > 0.5 AND BI-RADS_5 > 0.5 AND margin_? > 0.5 AND age_42 > 0.5 THEN CLASS=0
IF BI-RADS_4 <= 0.5 AND age_43 <= 0.5 AND age_42 <= 0.5 AND margin_3 <= 0.5 AND margin_? > 0.5 AND shape_3 > 0.5 THEN CLASS=0
IF shape_4 > 0.5 THEN CLASS=1
IF age_35 <= 0.5 AND shape_1 > 0.5 AND age_65 <= 0.5 AND margin_5 > 0.5 AND density_2 <= 0.5 THEN CLASS=1
IF age_35 <= 0.5 AND shape_1 > 0.5 AND age_65 > 0.5 AND margin_? > 0.5 AND BI-RADS_5 > 0.5 THEN CLASS=1
IF BI-RADS_5 > 0.5 AND shape_4 <= 0.5 AND margin_? <= 0.5 THEN CLASS=1
IF BI-RADS_3 <= 0.5 AND margin_? > 0.5 AND age_72 <= 0.5 AND shape_4 <= 0.5 AND age_55 > 0.5 AND shape_1 <= 0.5 AND 
density_3 <= 0.5 THEN CLASS=1
IF BI-RADS_4 > 0.161 AND age_50 <= 0.102 AND shape_1 > 0.747 AND density_2 > 0.565 THEN CLASS=0
IF age_27 <= 0.491 AND BI-RADS_5 > 0.095 AND shape_4 <= 0.362 AND age_60 <= 0.026 AND age_64 > 0.004 AND margin_? > 0.668 
THEN CLASS=0
IF BI-RADS_5 > 0.098 AND age_42 <= 0.514 AND density_3 > 0.78 AND margin_? > 0.504 AND age_63 <= 0.714 AND age_60 > 0.745 
THEN CLASS=0
IF shape_2 <= 0.402 AND shape_1 <= 0.123 AND margin_? <= 0.554 AND margin_5 > 0.17 AND BI-RADS_4 <= 0.605 AND age_67 <= 0.094 
AND shape_3 > 0.279 THEN CLASS=1
IF shape_4 <= 0.969 AND BI-RADS_5 <= 0.162 AND margin_5 <= 0.637 THEN CLASS=0
IF shape_1 > 0.25 AND age_79 > 0.953 AND density_2 <= 0.807 THEN CLASS=1
IF margin_5 > 0.5 AND age_65 <= 0.5 AND shape_? > 0.5 THEN CLASS=1
IF margin_4 <= 0.372 AND age_74 <= 0.976 AND shape_1 <= 0.028 AND BI-RADS_3 > 0.889 AND margin_5 > 0.88 THEN CLASS=1
IF shape_4 <= 0.974 AND margin_3 <= 0.403 AND age_36 > 0.583 THEN CLASS=0

\end{verbatim}
\end{tiny}
\caption{Explanation of the Mammographic mass data set.}
\label{explanation-mammographic-mass}
\end{figure}

\subsubsection{Blood Transfusion Service Center Data Set}

To demonstrate the RFMTC marketing model (a modified version of RFM), this study adopted the donor database of Blood Transfusion Service Center in Hsin-Chu City in Taiwan. The center passes their blood transfusion service bus to one university in Hsin-Chu City to gather blood donated about every three months. To build a FRMTC model, 748 donors were selected at random from the donor database. There are 4 attributes:
\begin{itemize}
\item R (Recency - months since last donation),
\item F (Frequency - total number of donation),
\item M (Monetary - total blood donated in c.c.),
\item T (Time - months since first donation), and
\end{itemize}
The class is a binary variable representing whether he/she donated blood in March 2007 (1 stand for donating blood; 0 stands for not donating blood).  The data set is described and can be downloaded here\footnote{\url{https://archive.ics.uci.edu/ml/datasets/Blood+Transfusion+Service+Center}}. The explanation is printed in Figure \ref{explanation-blood-transfusion}. The explanation is composed of rules, has a complexity of 76, a coverage of $100\%$ and has 150 classification errors.

\begin{figure}[H]
\begin{tiny}
\centering
\begin{verbatim}
IF R > 6.5 THEN CLASS=0
IF R <= 6.5 AND M > 1125.0 AND F <= 18.0 AND T > 49.5 THEN CLASS=0
IF R <= 6.5 AND M > 1125.0 AND F > 18.0 AND M <= 5625.0 THEN CLASS=1
IF R <= 6.5 AND M > 1125.0 AND M > 4500.0 AND M > 5625.0 AND F > 25.0 THEN CLASS=1
IF F > 4.5 AND R <= 6.5 AND F <= 18.0 AND T <= 49.5 AND T > 18.5 AND M > 1625.0 THEN CLASS=1
IF F > 4.5 AND R <= 6.5 AND M <= 10875.0 THEN CLASS=1
IF R <= 4.5 AND F > 4.5 AND T > 49.5 AND M <= 3125.0 AND F > 10.5 THEN CLASS=0
IF F <= 4.5 AND R > 6.5 AND M > 375.0 AND M > 875.0 AND T <= 24.5 AND R > 10.5 AND R > 15.0 THEN CLASS=0
IF F > 4.5 AND R <= 6.5 AND F <= 18.0 AND M > 2625.0 AND T <= 80.0 AND T <= 30.0 THEN CLASS=1
IF F > 4.5 AND R <= 6.5 AND T > 82.5 AND F <= 43.5 THEN CLASS=1
IF M <= 6164.961 AND F > 19.291 AND M > 5647.456 THEN CLASS=0
IF F <= 40.892 AND M > 2055.415 AND R > 11.469 AND R <= 15.171 THEN CLASS=0
IF F <= 32.328 AND M <= 2189.537 AND F > 5.305 AND T <= 20.136 AND F > 6.995 THEN CLASS=1
IF F <= 32.328 AND M <= 2189.537 AND F > 5.305 AND T > 20.136 AND R <= 2.249 AND M > 1716.326 AND T > 63.353 THEN CLASS=0
IF R > 6.5 AND F > 1.5 AND F > 17.5 THEN CLASS=0
IF F <= 25.768 AND T <= 63.91 AND R > 10.216 AND R <= 12.862 AND M > 2050.118 THEN CLASS=0
IF F <= 23.365 AND R <= 13.524 AND R <= 6.888 AND M <= 1579.136 THEN CLASS=0

\end{verbatim}
\end{tiny}
\caption{Explanation of the Blood Transfusion Service Center data set.}
\label{explanation-blood-transfusion}
\end{figure}

\subsubsection{Ecoli Data Set}

This data set contains information of Escherichia coli. It is a bacterium of the genus Escherichia that is commonly found in the lower intestine of warm-blooded organisms. The data set has 336 instances split into 8 classes. There are 8 attributes:
\begin{itemize}
    \item Sequence name: accession number for the SWISS-PROT database
    \item mcg: McGeoch's method for signal sequence recognition
    \item gvh: von Heijne's method for signal sequence recognition
    \item lip: von Heijne's Signal Peptidase II consensus sequence score (binary attribute)
    \item chg: presence of charge on N-terminus of predicted lipoproteins (binary attribute)
    \item aac: score of discriminant analysis of the amino acid content of outer membrane and periplasmic proteins
    \item alm1: score of the ALOM membrane spanning region prediction program
    \item alm2: score of ALOM program after excluding putative cleavable signal regions from the sequence
\end{itemize}
The data set is described and can be downloaded here\footnote{\url{https://archive.ics.uci.edu/ml/datasets/ecoli}}. The explanation is printed in Figure \ref{explanation-ecoli}. The explanation is composed of 29 rules, has a complexity of 112, a coverage of $99.4\%$ and 31 classification errors.

\begin{figure}[H]
\centering
\begin{tiny}
\begin{verbatim}
IF alm1 <= 0.575 AND gvh <= 0.585 AND mcg <= 0.62 THEN CLASS=cp
IF alm1 <= 0.575 AND gvh > 0.585 AND acc > 0.64 THEN CLASS=om
IF acc <= 0.565 AND alm1 <= 0.58 AND mcg <= 0.615 AND alm1 <= 0.365 AND gvh > 0.66 THEN CLASS=om
IF acc <= 0.565 AND alm1 <= 0.58 AND mcg > 0.615 AND gvh > 0.56 AND lip <= 0.74 THEN CLASS=pp
IF acc <= 0.565 AND alm1 <= 0.58 AND mcg > 0.615 AND gvh > 0.56 AND lip > 0.74 THEN CLASS=omL
IF acc <= 0.55 AND alm1 > 0.58 AND alm2 <= 0.77 AND mcg <= 0.62 THEN CLASS=im
IF acc > 0.64 AND alm2 <= 0.575 AND gvh > 0.48 AND alm1 <= 0.575 THEN CLASS=om
IF alm2 <= 0.615 AND alm1 <= 0.385 AND gvh <= 0.56 THEN CLASS=cp
IF alm1 <= 0.575 AND acc <= 0.64 AND gvh <= 0.585 AND chg > 0.75 THEN CLASS=imL
IF alm1 <= 0.575 AND acc <= 0.64 AND gvh > 0.585 AND acc <= 0.505 AND alm2 <= 0.375 THEN CLASS=pp
IF alm1 > 0.585 AND mcg > 0.615 AND lip <= 0.74 AND alm2 <= 0.59 THEN CLASS=pp
IF alm1 > 0.575 AND gvh <= 0.755 AND alm2 > 0.31 AND mcg <= 0.745 THEN CLASS=im
IF alm1 > 0.575 AND gvh <= 0.755 AND alm2 > 0.31 AND mcg > 0.745 THEN CLASS=imU
IF alm2 <= 0.615 AND gvh <= 0.565 AND alm1 <= 0.53 AND mcg > 0.71 AND gvh <= 0.54 THEN CLASS=imS
IF alm2 <= 0.615 AND gvh > 0.565 AND acc > 0.64 THEN CLASS=om
IF lip <= 0.956 AND alm2 > 0.678 AND mcg > 0.834 AND acc <= 0.585 AND acc > 0.539 THEN CLASS=imU
IF lip > 0.956 AND alm1 <= 0.679 AND acc <= 0.769 AND chg <= 0.963 THEN CLASS=omL
IF alm2 <= 0.645 AND acc <= 0.567 AND alm2 > 0.617 THEN CLASS=imU
IF alm2 <= 0.241 AND gvh > 0.574 AND alm2 > 0.161 AND gvh <= 0.73 AND alm2 > 0.239 THEN CLASS=pp
IF alm2 <= 0.241 AND gvh > 0.574 AND alm2 > 0.161 AND gvh > 0.73 THEN CLASS=pp
IF mcg > 0.745 AND alm1 > 0.6 AND mcg <= 0.82 THEN CLASS=imU
IF acc <= 0.495 AND lip <= 0.74 AND mcg <= 0.55 AND alm1 > 0.57 THEN CLASS=im
IF alm2 <= 0.615 AND gvh > 0.585 AND acc <= 0.64 AND alm1 <= 0.705 AND alm1 > 0.355 THEN CLASS=pp
IF alm2 <= 0.615 AND gvh > 0.585 AND acc > 0.64 THEN CLASS=om
IF alm2 > 0.835 AND alm1 > 0.845 AND acc <= 0.56 THEN CLASS=im
IF mcg <= 0.555 AND alm2 > 0.62 AND alm2 > 0.775 THEN CLASS=im
IF alm1 <= 0.355 AND alm2 > 0.385 THEN CLASS=cp
IF alm1 > 0.755 AND alm2 > 0.59 AND mcg <= 0.615 THEN CLASS=im
IF alm1 <= 0.575 AND alm2 > 0.385 AND mcg > 0.615 AND gvh > 0.52 THEN CLASS=pp

\end{verbatim}
\end{tiny}
\caption{Explanation of the Ecoli data set.}
\label{explanation-ecoli}
\end{figure}

\subsection{Tic-tac-toe Data Set}

This data set encodes the complete set of possible board configurations at the end of tic-tac-toe games, where \enquote{x} is assumed to have played first. The target concept is \enquote{win for x} (i.e., true when \enquote{x} has one of eight possible ways to create a "three-in-a-row"). The data set contains 958 instances split into 2 classes: \enquote{positive} and \enquote{negative}. There are 9 attributes.
\begin{itemize}
    \item top-left-square (TLS): \{x,o,b\}
    \item top-middle-square (TMS): \{x,o,b\}
    \item top-right-square (TRS): \{x,o,b\}
    \item middle-left-square (MLS): \{x,o,b\}
    \item middle-middle-square (MMS): \{x,o,b\}
    \item middle-right-square (MRS): \{x,o,b\}
    \item bottom-left-square (BLS): \{x,o,b\}
    \item bottom-middle-square (BMS): \{x,o,b\}
    \item bottom-right-square (BRS): \{x,o,b\}
\end{itemize}
The data set is described and can be downloaded here\footnote{\url{https://archive.ics.uci.edu/ml/datasets/Tic-Tac-Toe+Endgame}}. The explanation is printed in Figure \ref{explanation-tic-tac-toe}. It is composed of 31 rules, has a complexity of 117, a coverage of $85.1\%$ and 70 classification errors.

\begin{figure}[H]
\centering
\begin{tiny}
\begin{verbatim}
IF MMS_o <= 0.5 AND MMS_b > 0.5 AND BRS_x <= 0.5 AND TLS_x <= 0.5 THEN CLASS=negative
IF MMS_o <= 0.5 AND MMS_b > 0.5 AND BRS_x > 0.5 AND TLS_o <= 0.5 THEN CLASS=positive
IF MMS_o > 0.5 AND TRS_x <= 0.5 AND BLS_b > 0.5 THEN CLASS=negative
IF MMS_o > 0.5 AND BRS_x <= 0.5 AND TLS_b > 0.5 THEN CLASS=negative
IF MMS_o > 0.5 AND BRS_x > 0.5 AND TLS_o > 0.5 AND TMS_b > 0.5 THEN CLASS=positive
IF MLS_x <= 0.5 AND BLS_o <= 0.5 AND MMS_x > 0.5 AND TRS_o <= 0.5 THEN CLASS=positive
IF MLS_x <= 0.5 AND BLS_o > 0.5 AND TLS_o > 0.5 AND MLS_o > 0.5 THEN CLASS=negative
IF MLS_x > 0.5 AND TLS_x <= 0.5 AND BLS_o > 0.5 AND MMS_x > 0.5 THEN CLASS=positive
IF MLS_x > 0.5 AND TLS_x > 0.5 AND BLS_x > 0.5 THEN CLASS=positive
IF TRS_o <= 0.005 AND BRS_x <= 0.029 AND MMS_x <= 0.145 AND TLS_o > 0.558 THEN CLASS=negative
IF TRS_o <= 0.005 AND BRS_x > 0.029 AND MMS_b <= 0.331 AND MRS_x > 0.426 THEN CLASS=positive
IF TRS_o > 0.005 AND MMS_o > 0.243 AND BLS_x > 0.096 AND MRS_b > 0.097 THEN CLASS=positive
IF TLS_o <= 0.019 AND BRS_o <= 0.305 AND MMS_b > 0.55 THEN CLASS=positive
IF MMS_o <= 0.307 AND BLS_x <= 0.874 AND BRS_x > 0.223 AND TLS_o <= 0.218 THEN CLASS=positive
IF BRS_o <= 0.47 AND MMS_x > 0.813 AND TRS_b > 0.117 AND MLS_x > 0.962 THEN CLASS=positive
IF BRS_o > 0.47 AND MMS_x <= 0.372 AND MRS_x > 0.548 AND TLS_o > 0.968 THEN CLASS=negative
IF BRS_o > 0.47 AND MMS_x > 0.372 AND BLS_o <= 0.173 AND MRS_x > 0.244 THEN CLASS=positive
IF MMS_x <= 0.5 AND TRS_x <= 0.5 AND MRS_x > 0.5 AND BLS_o > 0.5 THEN CLASS=negative
IF MMS_x <= 0.5 AND TRS_x > 0.5 AND TLS_x > 0.5 AND MMS_o <= 0.5 THEN CLASS=positive
IF MMS_x <= 0.5 AND TRS_x > 0.5 AND TLS_x > 0.5 AND MMS_o > 0.5 THEN CLASS=positive
IF MMS_x > 0.5 AND BLS_b > 0.5 AND MLS_o > 0.5 AND TLS_o <= 0.5 THEN CLASS=positive
IF BRS_o <= 0.5 AND TLS_o <= 0.5 AND MMS_b <= 0.5 AND MMS_x > 0.5 THEN CLASS=positive
IF BRS_o > 0.5 AND TMS_b > 0.5 AND MLS_x <= 0.5 AND TRS_o > 0.5 THEN CLASS=negative
IF BMS_x <= 0.5 AND TRS_o > 0.5 AND TLS_x <= 0.5 AND MMS_o > 0.5 THEN CLASS=negative
IF MMS_x > 0.88 AND BRS_o <= 0.657 AND BLS_o <= 0.023 THEN CLASS=positive
IF MMS_o <= 0.85 AND BRS_o > 0.969 AND MMS_b > 0.435 AND TLS_b > 0.667 THEN CLASS=negative
IF MMS_o <= 0.585 AND TLS_x <= 0.386 AND TRS_o <= 0.94 AND BLS_b > 0.635 THEN CLASS=positive
IF MMS_o > 0.585 AND TLS_x <= 0.844 AND MLS_b > 0.903 AND MRS_o > 0.138 THEN CLASS=positive
IF BLS_o > 0.755 AND MMS_x <= 0.516 AND TRS_o > 0.809 THEN CLASS=negative
IF MMS_o <= 0.747 AND BLS_x > 0.726 AND TRS_o <= 0.282 THEN CLASS=positive
IF MMS_o > 0.747 AND BRS_b > 0.249 AND BMS_x > 0.063 AND TLS_x <= 0.968 THEN CLASS=negative

\end{verbatim}
\end{tiny}
\caption{Explanation of the Ecoli data set.}
\label{explanation-tic-tac-toe}
\end{figure}

\end{appendices}

\end{document}